\documentclass{article}
\usepackage{amsmath}
\usepackage{amssymb}
\usepackage{booktabs}
\usepackage{enumitem}
\usepackage{graphicx}
\usepackage{hyperref}
\usepackage{soul}
\usepackage{subcaption}
\usepackage[most]{tcolorbox}
\usepackage{wrapfig}
\usepackage{xcolor}
\definecolor{coral}{HTML}{FF6F61}
\definecolor{teal}{HTML}{009688}
\definecolor{sea_blue}{HTML}{006994}
\definecolor{prompt_frame}{HTML}{6A8EAE}
\definecolor{response_frame}{HTML}{E79A6A}
\definecolor{ground_truth_frame}{HTML}{8FBF8F}
\definecolor{summary_frame}{HTML}{D8A6E7}
\definecolor{comm_frame}{HTML}{FF8282}
\definecolor{highlight}{HTML}{CCEAE7}
\tcbuselibrary{breakable}

\usepackage[final]{corl_2025} 

\hypersetup{
    colorlinks,
    linkcolor=teal,
    citecolor=sea_blue,
    urlcolor=coral,
} 
\sethlcolor{highlight}
\graphicspath{{figs/}}

\title{CoRI: Communication of Robot Intent for\\Physical Human-Robot Interaction}

\author{
    Junxiang Wang\\
    Carnegie Mellon University\\
    \href{mailto:junxiang@cmu.edu}{\texttt{junxiang@cmu.edu}}\\
    \And
    Emek Barış Küçüktabak\\
    Honda Research Institute USA\\
    \href{mailto:baris_kucuktabak@honda-ri.com}{\texttt{baris\_kucuktabak@honda-ri.com}}\\
    \And
    Rana Soltani Zarrin\\
    Honda Research Institute USA\\
    \href{mailto:rana_soltanizarrin@honda-ri.com}{\texttt{rana\_soltanizarrin@honda-ri.com}}\\
    \And
    Zackory Erickson\\
    Carnegie Mellon University\\
    \href{mailto:zackory@cmu.edu}{\texttt{zackory@cmu.edu}}\\
}

\begin{document}
\maketitle


\begin{abstract}
    Clear communication of robot intent fosters transparency and interpretability in physical human-robot interaction (pHRI), particularly during assistive tasks involving direct human-robot contact. We introduce CoRI, a pipeline that automatically generates natural language communication of a robot's upcoming actions directly from its motion plan and visual perception. Our pipeline first processes the robot's image view to identify human poses and key environmental features. It then encodes the planned 3D spatial trajectory (including velocity and force) onto this view, visually grounding the path and its dynamics. CoRI queries a vision-language model with this visual representation to interpret the planned action within the visual context before generating concise, user-directed statements, without relying on task-specific information. Results from a user study involving robot-assisted feeding, bathing, and shaving tasks across two different robots indicate that CoRI leads to statistically significant difference in communication clarity compared to a baseline communication strategy. Specifically, CoRI effectively conveys not only the robot's high-level intentions but also crucial details about its motion and any collaborative user action needed. Video and code of our project can be found on our project website: \url{https://cori-phri.github.io/}.
\end{abstract}



\section{Introduction}
\label{sec:intro}

\begin{figure}[tbh]
    \centering
    \includegraphics[width=\linewidth]{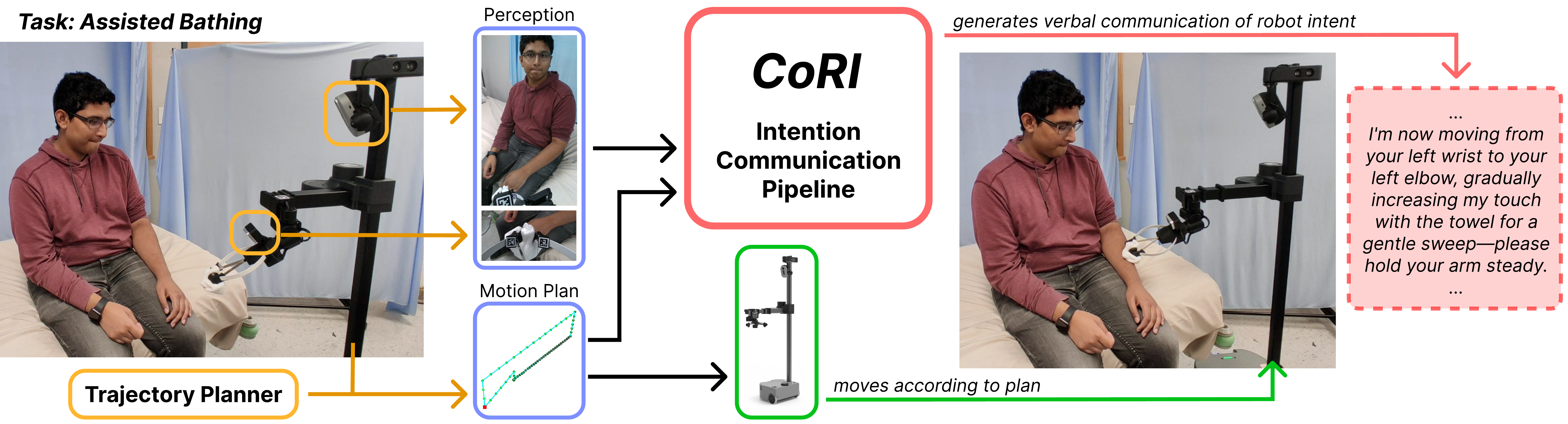}
    \caption{Our proposed pipeline, CoRI, generating intent communication during an assisted bathing task. Taking as inputs 1) visual perception from cameras and 2) motion plan from a trajectory planner, CoRI generates natural-language explanations of the robot's intention without requiring any task-specific knowledge. This explanation may then be spoken to the user as the robot executes its planned motion.}
    \label{fig:main}
\end{figure}

Robots are increasingly capable of interacting with humans physically, with wide applications in assistive scenarios, such as bathing, feeding, and dressing~\citep{nanavati2023physically}.
Without communication, the autonomous motions of a robot assisting with such tasks can be ambiguous and difficult to interpret, especially for non-experts. Thus, it is critical for robots to be able to clearly communicate their intentions using \emph{natural language}, which fosters transparency and interpretability~\citep{alhaji2021trust,mehrotra2024systematic,schott2023literature,st2015robot,fischer2018increasing}.
Compared to expert-oriented interfaces with steep learning curves that are currently used by robotics developers to interpret a robot's planned trajectories~\citep{berg2020review,li2024safe,huang2021survey}, natural language is immediately understandable even for first-time robot users.

In this paper, we propose \textbf{CoRI} (Communication of Robot Intent), a pipeline that allows any robot, regardless of its target task, to communicate to its user in the following three aspects:
\begin{itemize}[
    topsep=0pt,      
    partopsep=0pt,   
    parsep=0pt,      
    itemsep=0.1em,   
    leftmargin=2em,  
    labelsep=0.5em   
]
    \item The overall \textbf{intention} of the robot. This is the task-level goal that the robot aims to achieve.
    \item All aspects of the \textbf{motion} of the robot. This includes details regarding the planned trajectory itself---starting and ending positions, shape of the trajectory, as well as velocity and force profiles. Furthermore, these descriptions are intuitive---without any numerical values, but with references to the robot's surroundings and with comparative language.
    \item If needed, \textbf{user cooperation}. We recognize that many assistive tasks require certain behaviors from the user in order for the overall interaction to be successful (e.g. user taking a bite during feeding). Hence, our pipeline also conveys any desired human behavior.
\end{itemize}

As shown in Figure~\ref{fig:main}, CoRI only takes as inputs 1) the robot's image observation of its surroundings, and 2) its planned spatial waypoint trajectory, containing dynamics information such as position, velocity, and force.
These two inputs have drastically different representations---one a 2D image and the other a list of 3D waypoints---that must be jointly analyzed to infer the robot's intent if no task-specific information is provided.
With CoRI, we design a combined visual representation of the two inputs that captures various dynamics information, while emphasizing any interaction in the trajectory. We then use a vision-language model (VLM) to effectively interpret this visual representation, and subsequently a reasoning large-language model (LLM) to generate natural-language intent communication grounded in the environment context. CoRI is tested in three assistive tasks involving physical contact between the robot and the human user. We first generate intent communication with CoRI, and then narrate the generated sentences as the robot executes each of these tasks. We conduct a user study with these tasks on two different robots---Stretch, a mobile manipulator and xArm, a tabletop robot arm---and evaluate our pipeline against both a no-communication strategy and also a baseline strategy with scripted communication.

To summarize, we make the following contributions in this paper:
\begin{itemize}[
    topsep=0pt,      
    partopsep=0pt,   
    parsep=0pt,      
    itemsep=0.1em,   
    leftmargin=2em,  
    labelsep=0.5em   
]
    \item CoRI, the first task- and robot-agnostic pipeline for robots to communicate their intent to humans during physical interactions. It converts a robot's visual observation and its motion plan into environment-grounded natural-language statements that cover: 1) task-level intent, 2) motion dynamics, and 3) desired human cooperation.
    \item We design a set of operations to visually embed a robot's planned motion with dynamics information, and we show that even without any task-specific information, large vision-language models can reason on this visual representation to generate communications that are consistently aligned with the robot's ground-truth intent and motion descriptions.
    \item We conduct a user study with 16 participants on three physically assistive tasks (bathing, shaving, feeding) and two robot platforms, and we show that CoRI significantly improves users' understanding of the robot's intent compared to scripted or no-communication baselines.
\end{itemize}


\section{Related Work}
\label{sec:related-work}

\paragraph{Intention Communication in Robotics.}
Communicating a robot's intention can take various forms, including modalities beyond text and speech~\citep{pascher2023communicate}. One of the most popular approaches is to generate some form of visualization of the robot's goal or planned path. Some mechanisms include indicator lights onboard the robot~\citep{domonkos2020led,hetherington2021hey,lemasurier2021methods}, projection of the robot's next motions directly onto the world~\citep{hetherington2021hey,andersen2016projecting,wengefeld2020laser,chadalavada2020bi}, display screens for intent visualization~\citep{aubert2018designing,matsumaru2007mobile}, or augmented reality (AR) headsets worn by the human to observe the planned motion in 3D~\citep{walker2018communicating,gruenefeld2020mind,tsamis2021intuitive,gu2021seeing,rosen2019communicating}. Unlike these methods, our approach conveys rich intent information using natural language, making it adaptable across tasks and free from specialized visualization hardware.
Another line of work designs implicit communication through a robot's motion, either through gestures that carry semantic meaning~\citep{lemasurier2021methods,he2020investigating,mikawa2018expression,lohse2014robot,schreiter2023advantages} or specially-designed \emph{legible} motions, in the sense that the robot's goal would be clear to a person observing the robot's motion~\citep{dragan2013legibility,dragan2013generating,szafir2014communication,capelli2019communication}. In contrast, our approach communicates the intent behind any planned motion using language, without requiring specially crafted motions, and remains accessible even when visual cues are missed.
Explicit verbal communications in current literature sometimes appear in the form of scripted sentences~\citep{unhelkar2020decision,nikolaidis2018planning,lee2023effect} and may be paired with other forms of communication such as gestures~\citep{lohse2014robot,schreiter2023advantages,soderlund2023service}, but such scripted communication tend to see limited application areas. More free-form verbalization of a robot's plan mostly addresses the problem space of navigation~\citep{rosenthal2016verbalization,canal2021task}, where the robot's environment can be more easily mapped and represented semantically, compared to an arbitrary manipulation task. Our work is similar in the sense that we also generate purely verbal, unscripted communication. However, our method focuses on physical human-robot interaction, inferring high-level intent and desired user behavior from the robot's current state alone, without relying on prior knowledge of its morphology, task, or environment, making it more broadly applicable in various assistive scenarios.


\paragraph{VLMs and LLMs in Robotics.}
VLMs and LLMs are being employed successfully across a number of research thrusts in robotics, including reward function generation~\citep{yu2023language,ma2024eureka,xie2024textreward,wang2024rlvlmf}, planning~\citep{brohan2023can,huang2022language,singh2023progprompt}, and closed-loop low-level action generation~\citep{zitkovich2023rt,kim2024openvla,black2024pi}. Different from these applications listed, our work utilizes outputs of a LLM not to generate robot motions, but instead directly in the form of natural language as communication.
In this sense, some of the most relevant lines of work are the ones that embed LLMs in socially assistive robots and voice assistants for verbal communication, in the context of information retrieval, conversation, or issuing motion commands~\citep{kian2024can,mahmood2025user,padmanabha2024voicepilot}. Contrary from these systems that primarily respond to user-initiated interactions, our work enables robot-initiated communication to proactively convey intent during physically assistive tasks.
VLMs are also applied to facilitate communication in more collaborative HRI scenarios~\citep{wang2024mosaic,grannen2025provox,mandi2024roco}, where verbal communication is conductive to accomplish a common task goal. These works each focus on a single task domain, such as cooking, packing, and tabletop manipulation, and the communications from the robot to the human are operational, aimed at effectively achieving the task goal. In contrast, our pipeline generalizes across physically assistive tasks, and our communications are explanatory rather than operational, facilitating transparency and interpretability to help humans anticipate and understand the robot's intent.
Another line of work solves the problem of converting a robot's multimodal perception and action into natural language narration, mainly for the purpose of failure identification, explanation, and correction, directed at robot developers~\citep{liu2023reflect,wang2024can}. Although our work shares a similar idea of translating perception and action into language, we instead focus on communication of \emph{intent}, in language that is easy to understand for users with little technical experience.

\section{Problem Statement and Assumptions}
\label{sec:psa}

We consider the problem of generating natural-language intent communication for an arbitrary robot based on a planned motion that carries out some physically assistive task. We first make the assumption that there is a human within the robot's observation, as the robot's planned motion should involve interaction with a person. Next, the only two pieces of information available in synthesizing communication are the following (also shown in the top-left of Figure~\ref{fig:sys-chart}):
\begin{itemize}[
    topsep=0pt,      
    partopsep=0pt,   
    parsep=0pt,      
    itemsep=0.1em,   
    leftmargin=2em,  
    labelsep=0.5em   
]
    \item \textbf{Image} observation $o$---an RGB image of the environment and person from the robot's camera. This can come from a head-mounted camera in the case of a mobile robot, or a spatially-fixed camera in the case of a stationary robot arm. Additionally, if the robot is holding a tool, and the environment image cannot clearly capture the tool, then a wrist camera image $o_w$ may optionally be included for more details.
    \item \textbf{Trajectory} $\tau=\{x_i: x_i=(t_i,\mathbf{p}_i,g_i,v_i,f_i)\}_{i=1}^N$---a list of waypoints in 3D space. Each waypoint $x_i$ includes its timestamp $t_i\in\mathbb{R}^+$, the robot end-effector Cartesian position relative to the robot's base frame  $\mathbf{p}_i\in\mathbb{R}^3$ , open/closed status of the gripper $g_i\in\{0,1\}$, velocity magnitude $v_i\in\mathbb{R}^+$, and intended amount of force to be applied through the end-effector $f_i\in\mathbb{R}^+$. This trajectory can be generated from an arbitrary motion planner.
\end{itemize}
From these two pieces of information alone (i.e. without any knowledge on the specific task or robot), we would like to generate user-directed verbal communication that encapsulates the robot's intent, details on its motion, as well as any user cooperation needed (as described in \S~\ref{sec:intro}).
We additionally assume that if the trajectory were to be executed, then the robot's end-effector would be captured by the robot's camera for the majority of the motion (e.g. the end-effector can start slightly outside the frame, but all ``core'' motions should be in the frame).


\section{Method}
\label{sec:method}

\begin{figure}[tb]
    \centering
    \includegraphics[width=\linewidth]{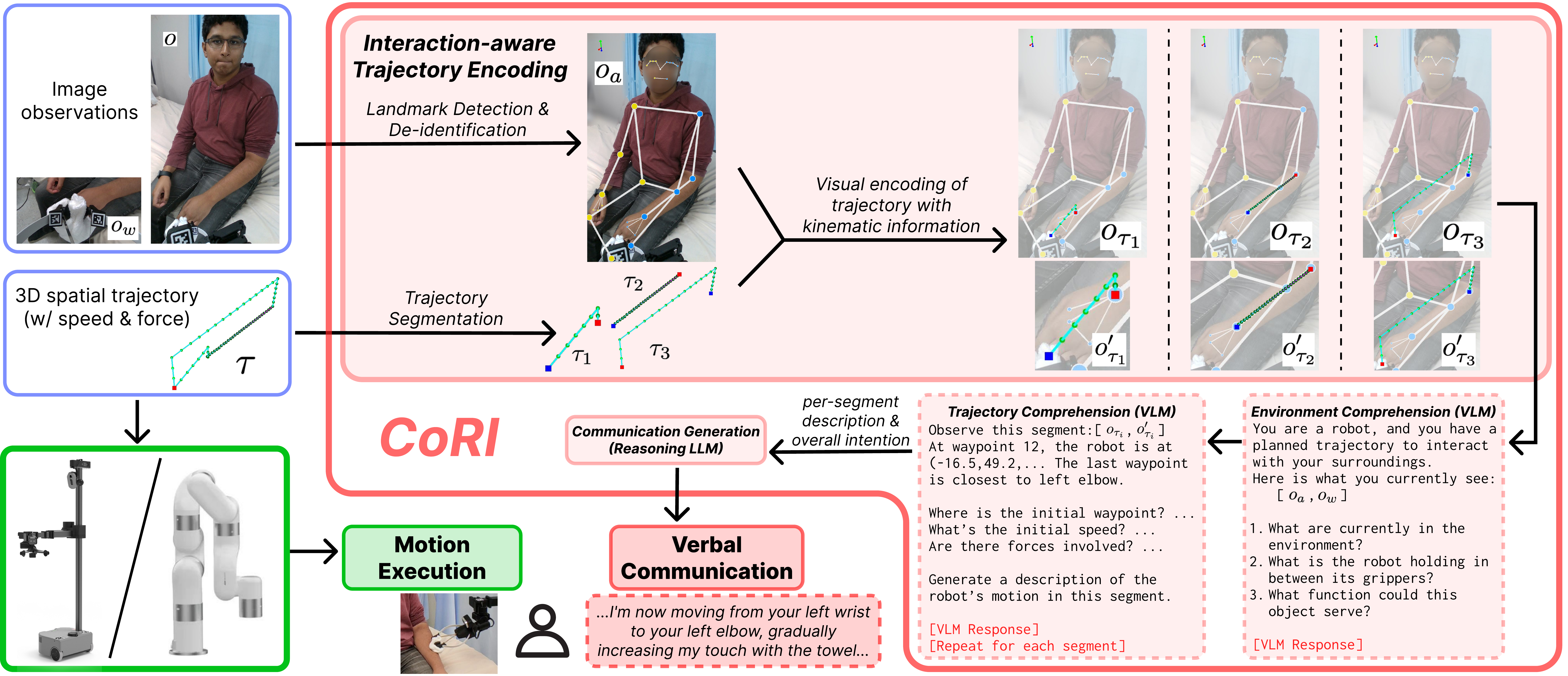}
    \caption{Overview of our CoRI pipeline. The pipeline takes as input an image observation $o$ of the environment and person, along with a planned 3D trajectory $\tau$. CoRI first performs \st{interaction-aware} trajectory encoding based on interaction to extract body landmarks, segment the trajectory, and visually encode the planned motion into visual context. Finally, it queries a VLM and a reasoning LLM respectively to interpret the visual information and generate user-oriented verbal communication.}
    \label{fig:sys-chart}
\end{figure}


Figure~\ref{fig:sys-chart} provides an overview of our CoRI pipeline, which consists of two stages. Our pipeline first visually encodes the trajectory $\tau$ in the context of the image observation $o$, including rich dynamics information while also recognizing  interaction events (\S~\ref{sec:traj-encoding}). This provides augmented visual context for subsequent visual reasoning queries to a VLM, before a reasoning LLM generates the final verbal communication (\S~\ref{sec:language-model}).

\subsection{Interaction-based Trajectory Encoding}
\label{sec:traj-encoding}



\paragraph{Body Landmark Detection and De-identification.}

Since we assume the image observation $o$ captures a person, CoRI first performs body pose recognition of the user through MediaPipe~\citep{lugaresi2019mediapipe}. Next, to preserve privacy when passing this image to a VLM, our pipeline blurs out the facial area of the user. The detected body landmarks are then annotated on the base image for enhanced spatial reasoning, including the facial landmarks (see top-middle of Figure~\ref{fig:sys-chart}). We denote this annotated and de-identified image as $o_a$.


\paragraph{Trajectory Segmentation.}

To facilitate clear interpretation and communication of robot intent, CoRI segments the trajectory $\tau$ based on meaningful interaction events, which we define based on the following three criteria:
\begin{enumerate}[
    topsep=-5pt,      
    partopsep=0pt,   
    parsep=0pt,      
    itemsep=0em,   
    leftmargin=2em,  
    labelsep=0.5em   
    ]
    \item Opening or closing of the gripper, indicating intended object retrieval.
    \item Presence of end-effector forces, indicating planned contact with a person or the environment.
    \item Pause ($>$2\,s) of the robot end-effector, which could be waiting for some user action.
\end{enumerate}
Therefore, CoRI breaks up the trajectory when there is a change in the gripper state, when there is a period of sustained nonzero external force, or when there is a pause. Mathematically, we represent the ordered set of indices $I\subset\{1,\dots,N-1\}$ at which the trajectory is segmented as follows:
\[I =\{i:\underbrace{(g_i\neq g_{i+1})}_{\text{gripper change}} \vee \underbrace{(f_i=0 \wedge f_{i+1}\neq0) \vee (f_i\neq0 \wedge f_{i+1}=0)}_{\text{onset or termination of force}} \vee \underbrace{(\mathbf{p}_i = \mathbf{p}_{i+1} \wedge t_{i+1}-t_i > 2)}_{\text{pause}}\}\]
where $\vee$ denote logical or and $\wedge$ denote logical and. If we denote the set of segmentation indices as $I=\{i_k\}_{k=1}^K$, the resulting set of trajectory segments can then be represented as:
\[\{\tau_1,\tau_2,\dots,\tau_{K+1}\} = \{\{x_1, x_2, \dots, x_{i_1}\}, \{x_{i_1+1}, x_{i_1+2}, \dots, x_{i_2}\}, \dots, \{x_{i_K+1}, x_{i_K+2}, \dots, x_N\}\}\]
This interaction-based segmentation allows each segment to be individually visualized and then interpreted by a VLM for greater clarity. We then mirror this segmentation in the final communication by generating one sentence per segment, enabling the user to receive the same structured understanding in verbal form. 


\paragraph{Visual Representation of Planned Trajectory.}
In order to ground the Cartesian trajectory in visual context for a VLM to reason, we project waypoints of each trajectory segment $\tau_i$ into the pixel space of the annotated image $o_a$. Our pipeline first creates a whitened version of the image annotated with body pose, and overlay these pixel-space waypoints on top (see top-right of Figure~\ref{fig:sys-chart}). The whitening process helps with better distinguishing the overlay from the actual world. The overlay is designed to contain visually-interpretable dynamics information---the starting waypoint in the segment is shown as a red square, the ending waypoint as a blue square, and all middle waypoints as circles colored in different shades of green, with darker green associated with slower velocity and brighter green higher velocity. Moreover, consecutive waypoints are connected with a straight line, which is colored according to the amount of force between the waypoints, mapped onto a color gradient from cyan to purple. We denote this overlay image per segment as $o_{\tau_i}$.
Recognizing that the overlay may only take up a small portion of the image $o_{\tau_i}$, we additionally create a cropped version $o'_{\tau_i}$ for each segment $\tau_i$, reduced to the area just around the trajectory in that segment (see top-right of Figure~\ref{fig:sys-chart}, row 2). With both the full overlay image $o_{\tau_i}$ and the cropped version $o'_{\tau_i}$ for each segment, the VLM can reason about both the high-level relationship between the trajectory segment and the person/environment, as well as the low-level dynamics of the segment itself.

\subsection{Language-model Querying}
\label{sec:language-model}



With the trajectory segmented and visually encoded, CoRI next interprets this information in context using a VLM in two stages, followed by generating natural language communication with a reasoning LLM. We use GPT-4o as the VLM and o3-mini as the reasoning LLM, with no fine-tuning performed. Examples of full prompts and responses can be found in the Appendix.

\paragraph{Environment Comprehension Stage.}
CoRI first queries the VLM to reason on the annotated image $o_a$, as well as the wrist camera image $o_w$ if provided. In this stage, the VLM is asked to output a description of $o_a$ and, if the gripper is closed at the start of the planned trajectory, a description of the object held by the robot and the object's potential function (see bottom-right of Figure~\ref{fig:sys-chart} for prompt snippets).

\paragraph{Trajectory Comprehension Stage.}

After the VLM responds to the previous query, CoRI next provides it with the overlay images $o_{\tau_k}$ and $o'_{\tau_k}$, one segment at a time. Alongside these images, all information from the trajectory segment $\tau_k$ is provided in textual form, as well as the closest body landmarks to the segment's start and end (see bottom-middle of Figure~\ref{fig:sys-chart} for examples).
In the same prompt with this visual and textual information, the VLM is asked simple questions that each targets one specific aspect of the motion. For example, ``Where is the blue square waypoint in the image?'' helps localize the starting point, and ``Are there forces involved? If so, in what region is this force being applied on?'' helps identify where contact happens. These questions are grouped by motion feature (position, velocity, or force), and most can be answered by observing the pair of images. Through these structured questions, the VLM incrementally builds an understanding of different aspects of the segment's motion, and at the end of the same prompt, the VLM is asked to generate a rich description of the segment, which incorporates the earlier answers.
For the last segment, CoRI additionally asks for an overall \emph{intention} for the entire trajectory $\tau$, as well as any actions required from the user to cooperate with the robot for each segment. Note that this intention reflects the task-level goal of the robot---distinct from the motions---and is deduced solely from reasoning about the trajectory $\tau$ within the observation $o$, without any prior knowledge about the underlying task. The desired user cooperation is likewise inferred through this visual reasoning. At the end of this stage, CoRI extracts the following textual outputs from the VLM: environment summary, per-segment summaries, overall intention, and any desired user cooperation.

\paragraph{Communication Generation Stage.}


With this textual information, CoRI queries the reasoning LLM to generate user-directed intent communications---spatially grounded and informative statements that are interpretable by the general populace, without jargon. As mentioned in \S~\ref{sec:traj-encoding}, CoRI outputs one sentence per trajectory segment, which communicates the overall intention, all aspects of motion, and desired user behavior (details as described in \S~\ref{sec:intro}). These statements can then be conveyed to a user over various modalities, and in this work, we specifically use speech.


\section{Experiments}
\label{sec:experiment}

To demonstrate that CoRI can effectively communicate robot intent and can be directly applied to different robot morphologies, we designed a series of physically assistive tasks on different robots (\S~\ref{sec:tasks}) and conducted a user study (\S~\ref{sec:user-study}) to obtain subjective evaluations from participants.

\subsection{Tasks and Robotic Systems}
\label{sec:tasks}

\begin{figure}[tb]
    \centering
    \includegraphics[width=\linewidth]{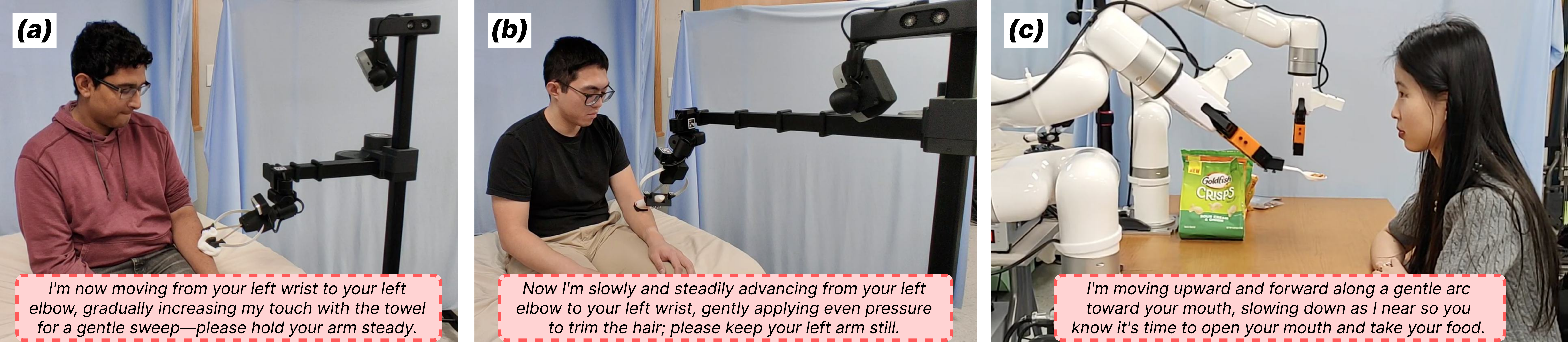}
    \caption{The three tasks implemented and used in the user study, along with example communications generated by CoRI: \textbf{(a)} simulated \emph{bathing}---moving a piece of dry washcloth along the user's forearm, \textbf{(b)} simulated \emph{shaving}---moving a surgical clipper with a fake razor along the user's forearm, and \textbf{(c)} \emph{feeding}---bringing the user a spoonful of food.}
    \label{fig:tasks}
\end{figure}

We designed motion planners for three different tasks on two robot setups to test the universality of our pipeline. Two of these tasks are performed by the Stretch 3 robot, a single-arm mobile manipulator. The first task is simulated \emph{bathing}, in which the robot guides a piece of dry washcloth on the user's arm. The second task is simulated \emph{shaving}, in which we replaced the razor on a surgical clipper with a 3D-printed plastic shaver head, and have the robot move it across the user's arm in different patterns to simulate trimming. The third task is \emph{feeding}, which is performed by an xArm 7 robot, a 7 degree-of-freedom manipulator mounted on a table. Figure~\ref{fig:tasks} shows example snapshots from the three tasks during one segment, along with CoRI's generated communication for that segment.

In each of the tasks, we designed two different trajectories for the planner, with differences intended to highlight the various types of information CoRI can communicate. The bathing task focuses on communicating changes in velocity and force, the shaving task on communicating the shape of the trajectory, and the feeding task on communicating desired human behavior. For example, in the bathing task, the first trajectory involves bathing the left forearm from wrist to elbow, in increasing force, and the second trajectory involves bathing the same region with constant force, but then continuing towards the shoulder at a higher velocity.

In terms of synchronizing the robot's motion with the communications, we first have the robot speak its sentence for the first segment $\tau_1$, and then start the robot's planned motion. For each of the subsequent segments, we time the robot's speech such that it first narrates its intent statement for the segment, and the corresponding movement only begins halfway into the speech. Symbolically, suppose the statement for segment $\tau_k$ would take a time of $t_k^s$ to be spoken verbally, then the statement starts playing at $0.5t_k^s$ before the motion in $\tau_k$ starts, i.e., the statement plays at $t_{i_{k-1}}-0.5t_k^s$ (recall $i_{k-1}$ is the waypoint index separating $\tau_{k-1}$ and $\tau_k$). In this way, each statement starts slightly ahead of the action it describes. Once the robot's motion starts, it does not pause specifically for communication; speech is embedded within continuous motion.

\subsection{User Study}
\label{sec:user-study}

We conducted a user study with 16 participants (7 female, 9 male, ages 18--49), each of whom experienced all three tasks, two different trajectories, and three different communication strategies---the CoRI method, a baseline scripted communication strategy commonly found in prior literature, and a no communication strategy---for a total of 18 trials per participant. The baseline communication strategy generates a per-segment statement using the body landmark closest to the end of the trajectory segment and saying ``I'm moving towards your \texttt{<LANDMARK NAME>}.'' This type of parameterizable scripted communication is commonly used as a baseline in other research on robot intent communication~\citep{andersen2016projecting,lohse2014robot,schreiter2023advantages}. The order of tasks and communication strategies was counterbalanced across participants. After each trial, the participants were asked to fill out a questionnaire of the following Likert items, on a 1-7 scale (7 for strongly agree, and 1 for strongly disagree):
\begin{enumerate}[
    topsep=-2pt,      
    partopsep=0pt,   
    parsep=0pt,      
    itemsep=0em,   
    leftmargin=4em,  
    labelsep=0.75em,   
    label=L\arabic*.
]
    \item I felt confident predicting the robot's next action.
    \item I understood what the robot was going to do.
    \item The robot's actions matched its communications.
    \item The robot correctly communicated its intentions.
    \item The robot's communications fully captured its actions in all notable aspects.
    \item The robot clearly communicated what I should do in the interaction.
\end{enumerate}
L1 and L2 are adapted from~\citep{mahadevan2024generative}, aiming to evaluate the user's level of understanding of the robot's motion, and the remaining questions are more directly targeted at the communication. L3 is to make sure that the communication is consistent with the motion, which we expect the baseline to also satisfy. L4--L6 specifically correspond to each of the three areas of desirable information covered in \S~\ref{sec:intro}: overall intention, all aspects of motion, and desired user behavior.


\section{Results and Discussion}
\label{sec:result}

\subsection{Likert Item Responses}
\label{sec:likert}

\begin{figure}[tb]
    \centering
    \includegraphics[width=\linewidth]{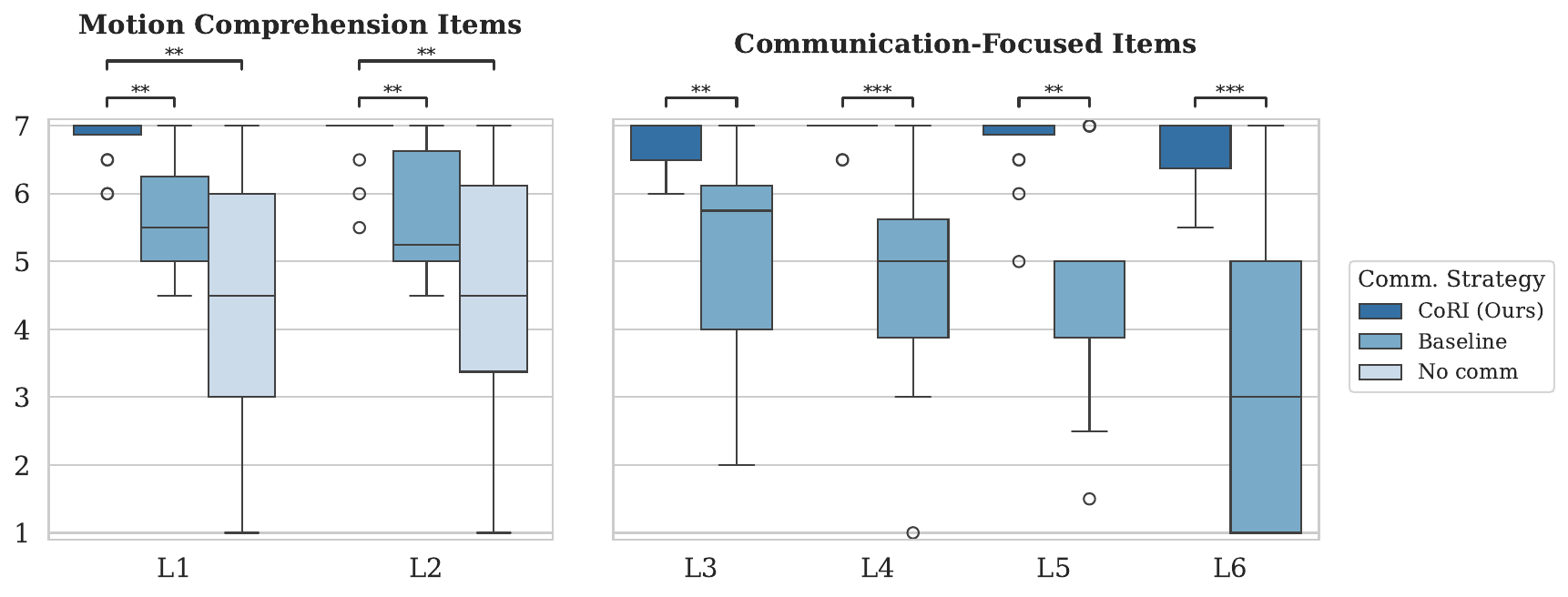}
    \caption{Box plots showing distribution of Likert-item responses for each participant, taking the median score per item and communication strategy. \textbf{Left}: Motion comprehension items (L1, L2), comparing our method (CoRI) against both the baseline and no-communication strategies. \textbf{Right}: Communication-related items (L3--L6), comparing CoRI against the baseline only. A Wilcoxon signed-rank test was used to assess statistical significance for each pairwise comparison involving CoRI. Asterisks denote significance levels ($p<0.05$, $p<0.01$, $p<0.001$).}
    \label{fig:box-plot}
\end{figure}

For each participant, we computed the median of their responses to each Likert question under each communication strategy. This yields one representative score per strategy for each participant and question. The resulting distributions are shown in Figure~\ref{fig:box-plot}. Since all participants experienced every communication strategy, we perform within-subject comparisons using pairwise Wilcoxon signed-rank tests. For motion comprehension items (L1, L2), we compare our method (CoRI) against both the scripted baseline and the no-communication strategy; for communication-focused items (L3--L6), we compare CoRI only against the baseline strategy, as the no-communication strategy doesn't apply here. All comparisons show statistically significant differences ($p<0.01$).

For both motion comprehension items (L1 and L2, left panel of Figure~\ref{fig:box-plot}), participants rated our method (CoRI) significantly higher than both the scripted baseline and the no-communication strategy ($p<0.01$ for all pairwise tests). Ratings for CoRI cluster near the maximum score of 7, indicating consistently better interpretation of robot motion compared to the baseline and no-communication strategies, which show greater variability and lower medians.

For communication-focused items (L3--L6; right panel of Figure~\ref{fig:box-plot}), CoRI also receives higher median scores than the baseline across all items. All pairwise tests show statistically significant differences ($p<0.01$ for all), with the largest differences in items L4 and L6 ($p<0.001$ for both), which specifically assess communicating task-level intention and eliciting appropriate user behavior, respectively. These two items require in-context trajectory reasoning, which CoRI achieves through visual trajectory encoding and interpretation through VLM. In contrast, the scripted baseline lacks contextual interpretation capabilities.


\subsection{Entailment within Ground Truth}
\label{sec:entailment}

In addition, for each trajectory segment, we design a ground-truth paragraph that contains information consistent with the intent and motion of the trajectory planner, including all dynamics features and possible collaborative user actions. These oracle descriptions are designed to encapsulate all correct information that could be communicated to the user, but too verbose if spoken verbally to the user directly. We then compare each of the statements generated by CoRI throughout the user study to the corresponding ground-truth paragraph, and obtain the probability that the generated statement is \emph{entailed} within the fixed ground-truth, through the RoBERTa-large model~\citep{liu2019roberta} fine-tuned on the Multi-Genre Natural Language Inference dataset (MNLI).
A statement then would receive a high entailment probability (values close to 1) only if the statement contains no information absent from the ground-truth paragraph, i.e. ``false'' information that is inconsistent with the robot's motion plan.
We additionally generate a one-sentence summary of each ground-truth paragraph through an LLM and obtain their entailment probability. These oracle summaries are not available to the robot at test time, but their entailment scores serve as approximate upper bounds to attainable scores. The ground-truth paragraphs and their summaries can be found in the Appendix.

\begin{table}[tbh]
    \vspace{-2mm}
    \centering
    \caption{Mean entailment probability (±1 SD). ``Oracle summaries (UB)'' refers to summaries of ground-truth paragraphs, which serve as approximate upper bounds for achievable scores.}
    \vspace{1mm}
    \begin{tabular}{lccc}
        \toprule
        Task & Bathing & Shaving & Feeding \\
        \midrule
        \textbf{CoRI (Ours)} & 0.95 (±0.03) & 0.95 (±0.06) & 0.95 (±0.08) \\
        Baseline & 0.88 (±0.21) & 0.70 (±0.40) & 0.69 (±0.31) \\
        Oracle summaries (UB) & 0.96 (±0.05) & 0.98 (±0.01) & 0.96 (±0.05) \\
        \bottomrule
    \end{tabular}
    \label{tab:entailment}
\end{table}


Table~\ref{tab:entailment} summarizes the mean entailment probability across all participants, grouped by task. As shown, the statements generated by CoRI achieve high entailment probability for all tasks and also score similarly to the oracle summaries, hence close to the best achievable score.
Overall, our results confirm that CoRI can reliably infer and communicate the robot's intent directly from images and planned trajectories, achieving near-oracle consistency without any explicit task information.


\section{Conclusion}
\label{sec:conclusion}

We present CoRI, a pipeline that converts a robot's visual perception and planned motion into user-directed natural language communication of the robot's intent for physical human-robot interaction. Through grounding the planned trajectory in visual context, we are able to generate statements that not only communicate the high-level intention, but also crucial aspects about the robot's motion as well as any desired human collaboration. We conducted a user study to evaluate the pipeline and observe that CoRI generates concise statements that are semantically similar to verbose oracle intent statements, while also outperforming baseline communication methods.


\clearpage
\section*{Limitations}
\label{sec:limitation}

\paragraph{Evaluation metrics.}
This work focuses on information completeness and accuracy of communications generated by CoRI, but does not consider psychological effects, such as perceived safety, comfort, or user trust. Further studies are needed to assess these psychological perceptions, which could be crucial to future widespread deployment of assistive robotic technologies.
\vspace{-3mm}
\paragraph{Task diversity.}
We observe strong results for three physically assistive human-robot interaction tasks on two robots; however, future work would benefit from evaluating how the intent communication pipeline extends to additional pHRI tasks---such as object handover, assisted dressing, and hair combing---as well as different robots.
\vspace{-3mm}
\paragraph{Closed-loop controllers and policies.}
CoRI currently supports preplanned trajectories and not closed-loop controllers or policies that generate short-horizon actions, trained using reinforcement learning or imitation learning. This is in part due to the latency associated with VLM and LLM queries. Additional investigation is needed to extend this framework to real-time control policies such as by leveraging local language models~\citep{touvron2023llama,deitke2024molmo} and through rolling out policies in a parallel simulation environment.
\vspace{-3mm}
\paragraph{Bidirectional Communication.}
CoRI accomplishes communication from robots to humans in assistive scenarios, but a smooth and natural interaction should also allow humans to communicate to robots, also in natural language, for modifying the robots' motions. This is part of our ongoing work.
\vspace{-3mm}
\paragraph{Human motion and disruptions during execution.}
CoRI assumes the user remains stationary throughout the robot's execution of its motion. In practice, small user movements can be accommodated by quickly re-planning the robot's trajectory---as long as the intended interaction does not change, the original communication remains valid, and the interaction will not be interrupted. However, if significant user motion requires a change in the robot's interaction strategy, CoRI would need to regenerate both the trajectory and the associated communication, which would disrupt the interaction. In general, CoRI is not capable of recovering from interruptions during robot execution that terminate the intended interaction.


\acknowledgments{The authors would like to thank members of Robotic Caregiving and Human Interaction (RCHI) Lab for their feedback throughout this work, especially Kavya Puthuveetil for their detailed and constructive suggestions. We also thank the participants of our user study for their time and insights. This work is supported by Honda Research Institute USA.}


\clearpage
\bibliography{references}  

\begin{thebibliography}{60}
\providecommand{\natexlab}[1]{#1}
\providecommand{\url}[1]{\texttt{#1}}
\expandafter\ifx\csname urlstyle\endcsname\relax
  \providecommand{\doi}[1]{doi: #1}\else
  \providecommand{\doi}{doi: \begingroup \urlstyle{rm}\Url}\fi

\bibitem[Nanavati et~al.(2023)Nanavati, Ranganeni, and Cakmak]{nanavati2023physically}
A.~Nanavati, V.~Ranganeni, and M.~Cakmak.
\newblock Physically assistive robots: A systematic review of mobile and manipulator robots that physically assist people with disabilities.
\newblock \emph{Annual Review of Control, Robotics, and Autonomous Systems}, 7, 2023.

\bibitem[Alhaji et~al.(2021)Alhaji, Prilla, and Rausch]{alhaji2021trust}
B.~Alhaji, M.~Prilla, and A.~Rausch.
\newblock Trust dynamics and verbal assurances in human robot physical collaboration.
\newblock \emph{Frontiers in Artificial Intelligence}, 4:\penalty0 703504, 2021.

\bibitem[Mehrotra et~al.(2024)Mehrotra, Degachi, Vereschak, Jonker, and Tielman]{mehrotra2024systematic}
S.~Mehrotra, C.~Degachi, O.~Vereschak, C.~M. Jonker, and M.~L. Tielman.
\newblock A systematic review on fostering appropriate trust in human-{AI} interaction: Trends, opportunities and challenges.
\newblock \emph{ACM Journal on Responsible Computing}, 1\penalty0 (4):\penalty0 1--45, 2024.

\bibitem[Sch{\"o}tt et~al.(2023)Sch{\"o}tt, Amin, and Butz]{schott2023literature}
S.~Y. Sch{\"o}tt, R.~M. Amin, and A.~Butz.
\newblock A literature survey of how to convey transparency in co-located human--robot interaction.
\newblock \emph{Multimodal Technologies and Interaction}, 7\penalty0 (3):\penalty0 25, 2023.

\bibitem[St.~Clair and Mataric(2015)]{st2015robot}
A.~St.~Clair and M.~Mataric.
\newblock How robot verbal feedback can improve team performance in human-robot task collaborations.
\newblock In \emph{Proceedings of the Tenth Annual ACM/IEEE International Conference on Human-Robot Interaction (HRI)}, pages 213--220, 2015.

\bibitem[Fischer et~al.(2018)Fischer, Weigelin, and Bodenhagen]{fischer2018increasing}
K.~Fischer, H.~M. Weigelin, and L.~Bodenhagen.
\newblock Increasing trust in human--robot medical interactions: effects of transparency and adaptability.
\newblock \emph{Paladyn, Journal of Behavioral Robotics}, 9\penalty0 (1):\penalty0 95--109, 2018.

\bibitem[Berg and Lu(2020)]{berg2020review}
J.~Berg and S.~Lu.
\newblock Review of interfaces for industrial human-robot interaction.
\newblock \emph{Current Robotics Reports}, 1\penalty0 (2):\penalty0 27--34, 2020.

\bibitem[Li et~al.(2024)Li, Hu, Zhou, and Pham]{li2024safe}
W.~Li, Y.~Hu, Y.~Zhou, and D.~T. Pham.
\newblock Safe human--robot collaboration for industrial settings: a survey.
\newblock \emph{Journal of Intelligent Manufacturing}, 35\penalty0 (5):\penalty0 2235--2261, 2024.

\bibitem[Huang et~al.(2021)Huang, Shen, Li, Fey, and Brecher]{huang2021survey}
Z.~Huang, Y.~Shen, J.~Li, M.~Fey, and C.~Brecher.
\newblock A survey on {AI}-driven digital twins in industry 4.0: Smart manufacturing and advanced robotics.
\newblock \emph{Sensors}, 21\penalty0 (19):\penalty0 6340, 2021.

\bibitem[Pascher et~al.(2023)Pascher, Gruenefeld, Schneegass, and Gerken]{pascher2023communicate}
M.~Pascher, U.~Gruenefeld, S.~Schneegass, and J.~Gerken.
\newblock How to communicate robot motion intent: A scoping review.
\newblock In \emph{Proceedings of the 2023 CHI Conference on Human Factors in Computing Systems}, pages 1--17, 2023.

\bibitem[Domonkos et~al.(2020)Domonkos, Dombi, and Botzheim]{domonkos2020led}
M.~Domonkos, Z.~Dombi, and J.~Botzheim.
\newblock Led strip based robot movement intention signs for human-robot interactions.
\newblock In \emph{Proceedings of the 2020 IEEE 20th International Symposium on Computational Intelligence and Informatics (CINTI)}, pages 121--126. IEEE, 2020.

\bibitem[Hetherington et~al.(2021)Hetherington, Croft, and Van~der Loos]{hetherington2021hey}
N.~J. Hetherington, E.~A. Croft, and H.~M. Van~der Loos.
\newblock Hey robot, which way are you going? nonverbal motion legibility cues for human-robot spatial interaction.
\newblock \emph{IEEE Robotics and Automation Letters}, 6\penalty0 (3):\penalty0 5010--5015, 2021.

\bibitem[Lemasurier et~al.(2021)Lemasurier, Bejerano, Albanese, Parrillo, Yanco, Amerson, Hetrick, and Phillips]{lemasurier2021methods}
G.~Lemasurier, G.~Bejerano, V.~Albanese, J.~Parrillo, H.~A. Yanco, N.~Amerson, R.~Hetrick, and E.~Phillips.
\newblock Methods for expressing robot intent for human--robot collaboration in shared workspaces.
\newblock \emph{ACM Transactions on Human-Robot Interaction (THRI)}, 10\penalty0 (4):\penalty0 1--27, 2021.

\bibitem[Andersen et~al.(2016)Andersen, Madsen, Moeslund, and Amor]{andersen2016projecting}
R.~S. Andersen, O.~Madsen, T.~B. Moeslund, and H.~B. Amor.
\newblock Projecting robot intentions into human environments.
\newblock In \emph{Proceedings of the 2016 25th IEEE International Symposium on Robot and Human Interactive Communication (RO-MAN)}, pages 294--301. IEEE, 2016.

\bibitem[Wengefeld et~al.(2020)Wengefeld, H{\"o}chemer, Lewandowski, K{\"o}hler, Beer, and Gross]{wengefeld2020laser}
T.~Wengefeld, D.~H{\"o}chemer, B.~Lewandowski, M.~K{\"o}hler, M.~Beer, and H.-M. Gross.
\newblock A laser projection system for robot intention communication and human robot interaction.
\newblock In \emph{Proceedings of the 2020 29th IEEE International Conference on Robot and Human Interactive Communication (RO-MAN)}, pages 259--265. IEEE, 2020.

\bibitem[Chadalavada et~al.(2020)Chadalavada, Andreasson, Schindler, Palm, and Lilienthal]{chadalavada2020bi}
R.~T. Chadalavada, H.~Andreasson, M.~Schindler, R.~Palm, and A.~J. Lilienthal.
\newblock Bi-directional navigation intent communication using spatial augmented reality and eye-tracking glasses for improved safety in human--robot interaction.
\newblock \emph{Robotics and Computer-Integrated Manufacturing}, 61:\penalty0 101830, 2020.

\bibitem[Aubert et~al.(2018)Aubert, Bader, and Hauser]{aubert2018designing}
M.~C. Aubert, H.~Bader, and K.~Hauser.
\newblock Designing multimodal intent communication strategies for conflict avoidance in industrial human-robot teams.
\newblock In \emph{Proceedings of the 2018 27th IEEE International Symposium on Robot and Human Interactive Communication (RO-MAN)}, pages 1018--1025, 2018.

\bibitem[Matsumaru(2007)]{matsumaru2007mobile}
T.~Matsumaru.
\newblock Mobile robot with preliminary-announcement and indication function of forthcoming operation using flat-panel display.
\newblock In \emph{Proceedings of the 2007 IEEE International Conference on Robotics and Automation (ICRA)}, pages 1774--1781. IEEE, 2007.

\bibitem[Walker et~al.(2018)Walker, Hedayati, Lee, and Szafir]{walker2018communicating}
M.~Walker, H.~Hedayati, J.~Lee, and D.~Szafir.
\newblock Communicating robot motion intent with augmented reality.
\newblock In \emph{Proceedings of the 2018 ACM/IEEE International Conference on Human-Robot Interaction (HRI)}, pages 316--324, 2018.

\bibitem[Gruenefeld et~al.(2020)Gruenefeld, Pr{\"a}del, Illing, Stratmann, Drolshagen, and Pfingsthorn]{gruenefeld2020mind}
U.~Gruenefeld, L.~Pr{\"a}del, J.~Illing, T.~Stratmann, S.~Drolshagen, and M.~Pfingsthorn.
\newblock Mind the {AR}m: realtime visualization of robot motion intent in head-mounted augmented reality.
\newblock In \emph{Proceedings of Mensch und Computer 2020}, pages 259--266, 2020.

\bibitem[Tsamis et~al.(2021)Tsamis, Chantziaras, Giakoumis, Kostavelis, Kargakos, Tsakiris, and Tzovaras]{tsamis2021intuitive}
G.~Tsamis, G.~Chantziaras, D.~Giakoumis, I.~Kostavelis, A.~Kargakos, A.~Tsakiris, and D.~Tzovaras.
\newblock Intuitive and safe interaction in multi-user human robot collaboration environments through augmented reality displays.
\newblock In \emph{Proceedings of the 2021 30th IEEE international conference on robot \& human interactive communication (RO-MAN)}, pages 520--526. IEEE, 2021.

\bibitem[Gu et~al.(2021)Gu, Cosgun, Chan, Drummond, and Croft]{gu2021seeing}
M.~Gu, A.~Cosgun, W.~P. Chan, T.~Drummond, and E.~Croft.
\newblock Seeing thru walls: Visualizing mobile robots in augmented reality.
\newblock In \emph{Proceedings of the 2021 30th IEEE International Conference on Robot \& Human Interactive Communication (RO-MAN)}, pages 406--411. IEEE, 2021.

\bibitem[Rosen et~al.(2019)Rosen, Whitney, Phillips, Chien, Tompkin, Konidaris, and Tellex]{rosen2019communicating}
E.~Rosen, D.~Whitney, E.~Phillips, G.~Chien, J.~Tompkin, G.~Konidaris, and S.~Tellex.
\newblock Communicating and controlling robot arm motion intent through mixed-reality head-mounted displays.
\newblock \emph{The International Journal of Robotics Research}, 38\penalty0 (12-13):\penalty0 1513--1526, 2019.

\bibitem[He et~al.(2020)He, van Maris, and Caleb-Solly]{he2020investigating}
J.~He, A.~van Maris, and P.~Caleb-Solly.
\newblock Investigating the effectiveness of different interaction modalities for spatial human-robot interaction.
\newblock In \emph{Companion of the 2020 ACM/IEEE International Conference on Human-Robot Interaction (HRI)}, pages 239--241, 2020.

\bibitem[Mikawa et~al.(2018)Mikawa, Yoshikawa, and Fujisawa]{mikawa2018expression}
M.~Mikawa, Y.~Yoshikawa, and M.~Fujisawa.
\newblock Expression of intention by rotational head movements for teleoperated mobile robot.
\newblock In \emph{Proceedings of the 2018 IEEE 15th International Workshop on Advanced Motion Control (AMC)}, pages 249--254. IEEE, 2018.

\bibitem[Lohse et~al.(2014)Lohse, Rothuis, Gallego-P{\'e}rez, Karreman, and Evers]{lohse2014robot}
M.~Lohse, R.~Rothuis, J.~Gallego-P{\'e}rez, D.~E. Karreman, and V.~Evers.
\newblock Robot gestures make difficult tasks easier: the impact of gestures on perceived workload and task performance.
\newblock In \emph{Proceedings of the SIGCHI conference on human factors in computing systems}, pages 1459--1466, 2014.

\bibitem[Schreiter et~al.(2023)Schreiter, Morillo-Mendez, Chadalavada, Rudenko, Billing, Magnusson, Arras, and Lilienthal]{schreiter2023advantages}
T.~Schreiter, L.~Morillo-Mendez, R.~T. Chadalavada, A.~Rudenko, E.~Billing, M.~Magnusson, K.~O. Arras, and A.~J. Lilienthal.
\newblock Advantages of multimodal versus verbal-only robot-to-human communication with an anthropomorphic robotic mock driver.
\newblock In \emph{Proceedings of the 2023 32nd IEEE International Conference on Robot and Human Interactive Communication (RO-MAN)}, pages 293--300. IEEE, 2023.

\bibitem[Dragan et~al.(2013)Dragan, Lee, and Srinivasa]{dragan2013legibility}
A.~D. Dragan, K.~C. Lee, and S.~S. Srinivasa.
\newblock Legibility and predictability of robot motion.
\newblock In \emph{Proceedings of the 2013 8th ACM/IEEE International Conference on Human-Robot Interaction (HRI)}, pages 301--308. IEEE, 2013.

\bibitem[Dragan and Srinivasa(2013)]{dragan2013generating}
A.~Dragan and S.~Srinivasa.
\newblock Generating legible motion.
\newblock In \emph{Proceedings of Robotics: Science and Systems}, June 2013.

\bibitem[Szafir et~al.(2014)Szafir, Mutlu, and Fong]{szafir2014communication}
D.~Szafir, B.~Mutlu, and T.~Fong.
\newblock Communication of intent in assistive free flyers.
\newblock In \emph{Proceedings of the 2014 9th ACM/IEEE International Conference on Human-Robot Interaction (HRI)}, pages 358--365, 2014.

\bibitem[Capelli et~al.(2019)Capelli, Secchi, and Sabattini]{capelli2019communication}
B.~Capelli, C.~Secchi, and L.~Sabattini.
\newblock Communication through motion: Legibility of multi-robot systems.
\newblock In \emph{Proceedings of the 2019 International Symposium on Multi-Robot and Multi-Agent Systems (MRS)}, pages 126--132. IEEE, 2019.

\bibitem[Unhelkar et~al.(2020)Unhelkar, Li, and Shah]{unhelkar2020decision}
V.~V. Unhelkar, S.~Li, and J.~A. Shah.
\newblock Decision-making for bidirectional communication in sequential human-robot collaborative tasks.
\newblock In \emph{Proceedings of the 2020 ACM/IEEE International Conference on Human-Robot Interaction}, pages 329--341, 2020.

\bibitem[Nikolaidis et~al.(2018)Nikolaidis, Kwon, Forlizzi, and Srinivasa]{nikolaidis2018planning}
S.~Nikolaidis, M.~Kwon, J.~Forlizzi, and S.~Srinivasa.
\newblock Planning with verbal communication for human-robot collaboration.
\newblock \emph{ACM Transactions on Human-Robot Interaction (THRI)}, 7\penalty0 (3):\penalty0 1--21, 2018.

\bibitem[Lee et~al.(2023)Lee, Krishna, Zaidi, Paleja, Chen, Hedlund-Botti, Schrum, and Gombolay]{lee2023effect}
K.~M. Lee, A.~Krishna, Z.~Zaidi, R.~Paleja, L.~Chen, E.~Hedlund-Botti, M.~Schrum, and M.~Gombolay.
\newblock The effect of robot skill level and communication in rapid, proximate human-robot collaboration.
\newblock In \emph{Proceedings of the 2023 ACM/IEEE International Conference on Human-Robot Interaction}, pages 261--270, 2023.

\bibitem[S{\"o}derlund(2023)]{soderlund2023service}
M.~S{\"o}derlund.
\newblock Service robot verbalization in service processes with moral implications and its impact on satisfaction.
\newblock \emph{Technological Forecasting and Social Change}, 196:\penalty0 122831, 2023.

\bibitem[Rosenthal et~al.(2016)Rosenthal, Selvaraj, and Veloso]{rosenthal2016verbalization}
S.~Rosenthal, S.~P. Selvaraj, and M.~Veloso.
\newblock Verbalization: narration of autonomous robot experience.
\newblock In \emph{Proceedings of the Twenty-Fifth International Joint Conference on Artificial Intelligence (IJCAI)}, pages 862--868, 2016.

\bibitem[Canal et~al.(2021)Canal, Krivic, Luff, and Coles]{canal2021task}
G.~Canal, S.~Krivic, P.~Luff, and A.~Coles.
\newblock Task plan verbalizations with causal justifications.
\newblock In \emph{ICAPS 2021 Workshop on Explainable AI Planning (XAIP)}, 2021.

\bibitem[Yu et~al.(2023)Yu, Gileadi, Fu, Kirmani, Lee, Arenas, Chiang, Erez, Hasenclever, Humplik, et~al.]{yu2023language}
W.~Yu, N.~Gileadi, C.~Fu, S.~Kirmani, K.-H. Lee, M.~G. Arenas, H.-T.~L. Chiang, T.~Erez, L.~Hasenclever, J.~Humplik, et~al.
\newblock Language to rewards for robotic skill synthesis.
\newblock In \emph{Proceedings of the 7th Conference on Robot Learning (CoRL)}, pages 374--404. PMLR, 2023.

\bibitem[Ma et~al.(2024)Ma, Liang, Wang, Huang, Bastani, Jayaraman, Zhu, Fan, and Anandkumar]{ma2024eureka}
Y.~J. Ma, W.~Liang, G.~Wang, D.-A. Huang, O.~Bastani, D.~Jayaraman, Y.~Zhu, L.~Fan, and A.~Anandkumar.
\newblock Eureka: Human-level reward design via coding large language models.
\newblock In \emph{Proceedings of the Twelfth International Conference on Learning Representations (ICLR)}, 2024.

\bibitem[Xie et~al.(2024)Xie, Zhao, Wu, Liu, Luo, Zhong, Yang, and Yu]{xie2024textreward}
T.~Xie, S.~Zhao, C.~H. Wu, Y.~Liu, Q.~Luo, V.~Zhong, Y.~Yang, and T.~Yu.
\newblock Text2reward: Reward shaping with language models for reinforcement learning.
\newblock In \emph{Proceedings of the Twelfth International Conference on Learning Representations (ICLR)}, 2024.

\bibitem[Wang et~al.(2024)Wang, Sun, Zhang, Xian, Biyik, Held, and Erickson]{wang2024rlvlmf}
Y.~Wang, Z.~Sun, J.~Zhang, Z.~Xian, E.~Biyik, D.~Held, and Z.~Erickson.
\newblock {RL}-{VLM}-{F}: Reinforcement learning from vision language foundation model feedback.
\newblock In \emph{Proceedings of the 41st International Conference on Machine Learning (ICML)}, 2024.

\bibitem[Brohan et~al.(2022)Brohan, Chebotar, Finn, Hausman, Herzog, Ho, Ibarz, Irpan, Jang, Julian, et~al.]{brohan2023can}
A.~Brohan, Y.~Chebotar, C.~Finn, K.~Hausman, A.~Herzog, D.~Ho, J.~Ibarz, A.~Irpan, E.~Jang, R.~Julian, et~al.
\newblock Do as {I} can, not as {I} say: Grounding language in robotic affordances.
\newblock In \emph{Proceedings of the 6th Conference on Robot Learning (CoRL)}, pages 287--318. PMLR, 2022.

\bibitem[Huang et~al.(2022)Huang, Abbeel, Pathak, and Mordatch]{huang2022language}
W.~Huang, P.~Abbeel, D.~Pathak, and I.~Mordatch.
\newblock Language models as zero-shot planners: Extracting actionable knowledge for embodied agents.
\newblock In \emph{Proceedings of the 39th International Conference on Machine Learning (ICML)}, pages 9118--9147. PMLR, 2022.

\bibitem[Singh et~al.(2023)Singh, Blukis, Mousavian, Goyal, Xu, Tremblay, Fox, Thomason, and Garg]{singh2023progprompt}
I.~Singh, V.~Blukis, A.~Mousavian, A.~Goyal, D.~Xu, J.~Tremblay, D.~Fox, J.~Thomason, and A.~Garg.
\newblock Prog{P}rompt: Generating situated robot task plans using large language models.
\newblock In \emph{2023 IEEE International Conference on Robotics and Automation (ICRA)}, pages 11523--11530. IEEE, 2023.

\bibitem[Zitkovich et~al.(2023)Zitkovich, Yu, Xu, Xu, Xiao, Xia, Wu, Wohlhart, Welker, Wahid, et~al.]{zitkovich2023rt}
B.~Zitkovich, T.~Yu, S.~Xu, P.~Xu, T.~Xiao, F.~Xia, J.~Wu, P.~Wohlhart, S.~Welker, A.~Wahid, et~al.
\newblock {RT}-2: Vision-language-action models transfer web knowledge to robotic control.
\newblock In \emph{Proceedings of the 7th Conference on Robot Learning (CoRL)}, pages 2165--2183. PMLR, 2023.

\bibitem[Kim et~al.(2024)Kim, Pertsch, Karamcheti, Xiao, Balakrishna, Nair, Rafailov, Foster, Sanketi, Vuong, Kollar, Burchfiel, Tedrake, Sadigh, Levine, Liang, and Finn]{kim2024openvla}
M.~J. Kim, K.~Pertsch, S.~Karamcheti, T.~Xiao, A.~Balakrishna, S.~Nair, R.~Rafailov, E.~P. Foster, P.~R. Sanketi, Q.~Vuong, T.~Kollar, B.~Burchfiel, R.~Tedrake, D.~Sadigh, S.~Levine, P.~Liang, and C.~Finn.
\newblock Open{VLA}: An open-source vision-language-action model.
\newblock In \emph{Proceedings of the 8th Conference on Robot Learning (CoRL)}. PMLR, 2024.

\bibitem[Black et~al.(2024)Black, Brown, Driess, Esmail, Equi, Finn, Fusai, Groom, Hausman, Ichter, et~al.]{black2024pi}
K.~Black, N.~Brown, D.~Driess, A.~Esmail, M.~Equi, C.~Finn, N.~Fusai, L.~Groom, K.~Hausman, B.~Ichter, et~al.
\newblock $\pi_0 $: A vision-language-action flow model for general robot control.
\newblock \emph{arXiv preprint arXiv:2410.24164}, 2024.

\bibitem[Kian et~al.(2024)Kian, Zong, Fischer, Singh, Velentza, Sang, Upadhyay, Gupta, Faruki, Browning, et~al.]{kian2024can}
M.~J. Kian, M.~Zong, K.~Fischer, A.~Singh, A.-M. Velentza, P.~Sang, S.~Upadhyay, A.~Gupta, M.~A. Faruki, W.~Browning, et~al.
\newblock Can an {LLM}-powered socially assistive robot effectively and safely deliver cognitive behavioral therapy? a study with university students.
\newblock \emph{arXiv preprint arXiv:2402.17937}, 2024.

\bibitem[Mahmood et~al.(2025)Mahmood, Wang, Yao, Wang, and Huang]{mahmood2025user}
A.~Mahmood, J.~Wang, B.~Yao, D.~Wang, and C.-M. Huang.
\newblock User interaction patterns and breakdowns in conversing with {LLM}-powered voice assistants.
\newblock \emph{International Journal of Human-Computer Studies}, 195:\penalty0 103406, 2025.

\bibitem[Padmanabha et~al.(2024)Padmanabha, Yuan, Gupta, Karachiwalla, Majidi, Admoni, and Erickson]{padmanabha2024voicepilot}
A.~Padmanabha, J.~Yuan, J.~Gupta, Z.~Karachiwalla, C.~Majidi, H.~Admoni, and Z.~Erickson.
\newblock Voicepilot: Harnessing {LLM}s as speech interfaces for physically assistive robots.
\newblock In \emph{Proceedings of the 37th Annual ACM Symposium on User Interface Software and Technology}, pages 1--18, 2024.

\bibitem[Wang et~al.(2024)Wang, Kedia, Ren, Abdullah, Bhardwaj, Chao, Chen, Chin, Dan, Fan, Gonzalez-Pumariega, Kompella, Pace, Sharma, Sun, Sunkara, and Choudhury]{wang2024mosaic}
H.~Wang, K.~Kedia, J.~Ren, R.~Abdullah, A.~Bhardwaj, A.~Chao, K.~Y. Chen, N.~Chin, P.~Dan, X.~Fan, G.~Gonzalez-Pumariega, A.~Kompella, M.~A. Pace, Y.~Sharma, X.~Sun, N.~Sunkara, and S.~Choudhury.
\newblock Mosaic: A modular system for assistive and interactive cooking.
\newblock In \emph{Proceedings of the 8th Conference on Robot Learning (CoRL)}. PMLR, 2024.

\bibitem[Grannen et~al.(2025)Grannen, Karamcheti, Wulfe, and Sadigh]{grannen2025provox}
J.~Grannen, S.~Karamcheti, B.~Wulfe, and D.~Sadigh.
\newblock Provox: Personalization and proactive planning for situated human-robot collaboration.
\newblock \emph{arXiv preprint arXiv:2506.12248}, 2025.

\bibitem[Mandi et~al.(2024)Mandi, Jain, and Song]{mandi2024roco}
Z.~Mandi, S.~Jain, and S.~Song.
\newblock Roco: Dialectic multi-robot collaboration with large language models.
\newblock In \emph{2024 IEEE International Conference on Robotics and Automation (ICRA)}, pages 286--299. IEEE, 2024.

\bibitem[Liu et~al.(2023)Liu, Bahety, and Song]{liu2023reflect}
Z.~Liu, A.~Bahety, and S.~Song.
\newblock {REFLECT}: Summarizing robot experiences for failure explanation and correction.
\newblock In \emph{Proceedings of the 7th Conference on Robot Learning (CoRL)}, pages 3468--3484. PMLR, 2023.

\bibitem[Wang et~al.(2024)Wang, Liang, Dhat, Brumbaugh, Walker, Krishna, and Cakmak]{wang2024can}
Z.~Wang, B.~Liang, V.~Dhat, Z.~Brumbaugh, N.~Walker, R.~Krishna, and M.~Cakmak.
\newblock I can tell what i am doing: Toward real-world natural language grounding of robot experiences.
\newblock In \emph{Proceedings of the 8th Conference on Robot Learning (CoRL)}. PMLR, 2024.

\bibitem[Lugaresi et~al.(2019)Lugaresi, Tang, Nash, McClanahan, Uboweja, Hays, Zhang, Chang, Yong, Lee, et~al.]{lugaresi2019mediapipe}
C.~Lugaresi, J.~Tang, H.~Nash, C.~McClanahan, E.~Uboweja, M.~Hays, F.~Zhang, C.-L. Chang, M.~Yong, J.~Lee, et~al.
\newblock Media{P}ipe: A framework for perceiving and processing reality.
\newblock In \emph{Third Workshop on Computer Vision for AR/VR at IEEE Computer Vision and Pattern Recognition (CVPR)}, 2019.

\bibitem[Mahadevan et~al.(2024)Mahadevan, Chien, Brown, Xu, Parada, Xia, Zeng, Takayama, and Sadigh]{mahadevan2024generative}
K.~Mahadevan, J.~Chien, N.~Brown, Z.~Xu, C.~Parada, F.~Xia, A.~Zeng, L.~Takayama, and D.~Sadigh.
\newblock Generative expressive robot behaviors using large language models.
\newblock In \emph{Proceedings of the 2024 ACM/IEEE International Conference on Human-Robot Interaction (HRI)}, pages 482--491, 2024.

\bibitem[Liu et~al.(2019)Liu, Ott, Goyal, Du, Joshi, Chen, Levy, Lewis, Zettlemoyer, and Stoyanov]{liu2019roberta}
Y.~Liu, M.~Ott, N.~Goyal, J.~Du, M.~Joshi, D.~Chen, O.~Levy, M.~Lewis, L.~Zettlemoyer, and V.~Stoyanov.
\newblock Ro{BERT}a: A robustly optimized {BERT} pretraining approach.
\newblock \emph{arXiv preprint arXiv:1907.11692}, 2019.

\bibitem[Touvron et~al.(2023)Touvron, Lavril, Izacard, Martinet, Lachaux, Lacroix, Rozi{\`e}re, Goyal, Hambro, Azhar, et~al.]{touvron2023llama}
H.~Touvron, T.~Lavril, G.~Izacard, X.~Martinet, M.-A. Lachaux, T.~Lacroix, B.~Rozi{\`e}re, N.~Goyal, E.~Hambro, F.~Azhar, et~al.
\newblock Llama: Open and efficient foundation language models.
\newblock \emph{arXiv preprint arXiv:2302.13971}, 2023.

\bibitem[Deitke et~al.(2024)Deitke, Clark, Lee, Tripathi, Yang, Park, Salehi, Muennighoff, Lo, Soldaini, et~al.]{deitke2024molmo}
M.~Deitke, C.~Clark, S.~Lee, R.~Tripathi, Y.~Yang, J.~S. Park, M.~Salehi, N.~Muennighoff, K.~Lo, L.~Soldaini, et~al.
\newblock Molmo and {P}ix{M}o: Open weights and open data for state-of-the-art multimodal models.
\newblock \emph{arXiv preprint arXiv:2409.17146}, 2024.

\end{thebibliography}
\clearpage
\appendix
\section{Example Prompts and Responses}

Given each pair of inputs (image $o$ and trajectory $\tau$), CoRI makes queries to large vision-language models in different stages in order to generate intent communication (\S~\ref{sec:language-model}). In each prompt, most of the content is static---fixed text that is not dependent on the inputs. Only certain sections of the prompts are dynamically determined based on the inputs, especially on the trajectory $\tau$. Below we show example prompts and responses corresponding to the particular trial of the bathing task shown in Figure~\ref{fig:sys-chart} of the paper. We color all input-dependent, dynamically-determined information in {\color{prompt_frame!125}navy}.

\subsection{Environment Comprehension Stage}

In the prompt for the environment comprehension stage, we query the VLM with two input images (annotated image $o_a$ and wrist camera image $o_w$) and ask for a description of the robot's surroundings in open-ended fashion. Additionally, only if the gripper is closed at the start of the trajectory---i.e. if $g_1=0$---we also ask the VLM to give a description of the object held by the gripper, and its potential function (relevant questions colored in {\color{prompt_frame!125}navy}) in the following prompt example.
Note that the wrist camera image $o_w$ is only to better observe any object the robot is holding. If the robot is not holding any object, or if $o_a$ can already capture the object well enough, then only $o_a$ is needed for the reasoning, and the wording of the prompt can be easily modified to accommodate for the change. 

\begin{tcolorbox}[title=Prompt, breakable,
    colback=prompt_frame!10, colframe=prompt_frame!50,
    left=2mm, right=2mm,       
    top=1mm, bottom=1mm]
\# General descriptions:\\
You are a robot, where you have planned a trajectory to interact with what you see around you. You have a gripper that may hold an object.\\
You are given two base images of what you currently see, with the first one being a big-picture view, and the second one from a camera mounted at the wrist of the gripper, capturing the content held by the gripper. For privacy purposes, all humans being captured in the picture have their face blurred out. However, all major body joints have been detected and labeled, as well as notable facial landmarks (eyes, nose, and mouth). Note that a light blue-colored landmark means it's the person's left side, and a yellow-colored landmark means the person's right side. Remember this in your reasoning. Observe, and respond to the following questions:\\
What are currently in the environment?\\
{\color{prompt_frame!125}\emph{[The following questions are added only if the gripper is closed at the start of the trajectory.]}}\\
{\color{prompt_frame!125}What is the robot holding in between its grippers?\\
Use your most likely guess for what the thing is, and don't infer beyond what you see. Suppose the robot does not let go of what it's holding. Then what function could this thing serve, in the context?}
\end{tcolorbox}

\begin{tcolorbox}[title=Response, breakable,
    colback=response_frame!10, colframe=response_frame!50,
    left=2mm, right=2mm,       
    top=1mm, bottom=1mm]
\#\#\# Environment:\\
- A person is sitting on a bed or couch, wearing a maroon hoodie and jeans.\\
- The person's body joints and facial landmarks are labeled with blue and yellow markers, indicating left and right sides.\\
- The setting appears to be indoors, possibly a room or living area.\\

\#\#\# Object in Gripper:\\
- The robot is holding a white cloth or towel between its grippers.\\

\#\#\# Possible Function:\\
- If the robot does not let go of the towel, it could be used for cleaning or wiping purposes, such as assisting the person with personal hygiene or tidying up the surrounding area.
\end{tcolorbox}

\subsection{Trajectory Comprehension Stage}

Recall that CoRI makes a query for each trajectory segment $\tau_k$ in this stage, where the image inputs are $o_{\tau_k}$ and $o'_{\tau_k}$ for segment $k$. Each query consists of three sections: ``Understanding your planned trajectory'', ``Kinematic description for image $k$'', and lastly ``Task descriptions''. The first section describes the visual trajectory, and also gives a description of the textual trajectory information, which is contained in the second section. The last section lists the series of questions for the VLM to answer, as well as asking for the overall description. Therefore, only the middle section---the section on textual trajectory information---changes across each segment. The other two sections are completely fixed. Therefore, we only show these fixed sections for the first segment for simplicity. Extended numerical information is also omitted for clarity. Note that in the prompt, ``image'' is used interchangeably with ``segment''.

\begin{tcolorbox}[title=Prompt ($\tau_1$), breakable,
    colback=prompt_frame!10, colframe=prompt_frame!50,
    left=2mm, right=2mm,       
    top=1mm, bottom=1mm]
\# \textbf{Understanding your planned trajectory}\\
Your planned trajectory is represented both visually through images and textually. Below are explanations for the meaning of these representations.\\
\#\# Planned trajectory represented in images:\\
Here is sequence of images showing your planned trajectory as overlays on the base image. You will be shown the sequence one at a time, each accompanied by a cropped version that better displays the trajectory. Your planned trajectory is shown as waypoints connected by lines. In each image, the starting position is shown as a blue square, and the ending position as a red square. The rest of the waypoints in the middle are shown as circles in different shades of green, darker with lower velocities, and brighter with higher velocities. As another visual cue beyond the color, the length of the line segment connecting two consecutive waypoints is directly proportional to the speed at which the segment is travelled at. Additionally, forces are also represented in the images: The line segments connecting the waypoints are shown in cyan if the segment doesn't have any force. If there is force, then the line segment is colored in a gradient that shows cyan when the force is small and magenta when the force is large. These small and large are relative to the segment itself, and not absolute. Each subsequent image in the sequence starts at the last waypoint of the previous image and continues the trajectory. The base image is whitened for you to more clearly observe the overlay trajectories. \\
\#\# Overview of textual description to the planned trajectory:\\
For each image in the sequence, there is textual description of the robot end effector's position, velocity, force, and other relevant information, mostly regarding the waypoints shown in the images. The meaning of these descriptions are explained below: \\
\#\#\# Position descriptions:\\
These describe the position of the robot end-effector at each waypoint, as well as the relative translation between adjacent waypoints. Units are in centimeters.\\
\#\#\# Velocity descriptions:\\
These describe the average velocity of the robot end-effector as it goes between adjacent waypoints. Units are in centimeters per second.\\
\#\#\# Force descriptions:\\
These describe the force profile of the robot end-effector as it goes between adjacent waypoints, if there are moments of nonzero forces between two waypoints. Each nonzero profile is represented as a list, showing the change in force from one waypoint to the next. The first entry in the list represents the force at the starting waypoint, and the last entry represents the force at the ending waypoint. The middle entries represent the force profile as the robot end-effector moves from the starting waypoint to the ending waypoint. These forces represent the amount of force that the robot's end-effector is applying to some external object, and ignore the effect of anything that the robot is holding. This means that the robot plans to be in contact with some external object if and only if there are forces present. Units are in Newtons.\\
\#\#\# Coordinate convention\\
The coordinate system in these descriptions are defined relative to the robot's base, and an illustration of this convention is shown in the top left corner of every image, where red represents the positive x direction, green represents the positive y direction, and blue represents the positive z direction. The positive z direction is also the 'upwards' direction in real world, so it is upwards relative to any reference object that typically lies in the x-y plane (e.g. tables, floors).\\
\#\# Human body landmark positions\\
In addition to the robot planned trajectory, you will also be told the position of each of the visible body landmarks of any human being captured in the image. For example, if an arm is visible, then you will be told the positions of the wrist, elbow, and shoulder. These positions are defined in the same coordinate system as the trajectory. Additionally, for each image, you will be told which landmark is closest to the starting and ending waypoint of the trajectory respectively, along with the distance to the waypoint.\\
\#\#\# Gripper status\\
You will be told the gripper state (open or closed) at the start of each image. It maintains the same throughout the same image, and may only change in between images. Note that the gripper is closed at the start, and whatever the robot is holding will continue to be held, unless there is further communication saying that the gripper has been opened. Therefore, while the gripper is holding the object, reason about the function of the object and how it might interact with the environment.\\

\# \textbf{Kinematic description for image 1}\\
At the start of image {\color{prompt_frame!125}1}, the gripper is {\color{prompt_frame!125}closed}.\\
\#\# Position descriptions\\
At waypoint {\color{prompt_frame!125}1}, the position of the gripper center is at {\color{prompt_frame!125}(-4.3, -39.1, 79.3)}. From waypoint {\color{prompt_frame!125}1} to waypoint {\color{prompt_frame!125}2}, the general motion of the robot is {\color{prompt_frame!125}(-2.0, -1.7, -0.2)} centimeters. \texttt{[...]} At waypoint {\color{prompt_frame!125}9}, the position of the gripper center is at {\color{prompt_frame!125}(-16.0, -49.2, 74.2)}. \\
\#\# Velocity descriptions\\
From waypoint {\color{prompt_frame!125}1} to waypoint {\color{prompt_frame!125}2}, the average velocity of the robot is {\color{prompt_frame!125}3.0} centimeters per second. \texttt{[...]} From waypoint {\color{prompt_frame!125}8} to waypoint {\color{prompt_frame!125}9}, the average velocity of the robot is {\color{prompt_frame!125}2.0} centimeters per second. \\
\#\# Force descriptions\\
{\color{prompt_frame!125}There is no external force planned during this section.} \\
\#\# Human body landmark positions\\
The following are the positions for each of the detected body landmarks in the image: {\color{prompt_frame!125}left wrist: (-16.0, -49.2, 74.2) centimeters, \texttt{[...]}}\\
The first waypoint in this segment is closest to the {\color{prompt_frame!125}left wrist} at a distance of {\color{prompt_frame!125}16.3} centimeters. The last waypoint in this segment is closest to the {\color{prompt_frame!125}left wrist} at a distance of {\color{prompt_frame!125}0.0} centimeters. \\

\# \textbf{Task descriptions}\\
Your high-level goal is to summarize what the planned trajectory of the robot is trying to accomplish, and do so without numerical values, but with references to other components of the environment. You will answer a series of questions as you move towards building a complete statement.\\
For each image (segment), you answer the following questions:\\
{\setlength{\parindent}{1.5em}
1) Consider the position of the robot end-effector. Answer the following primarily from looking at the images, but refer to human body landmarks in the exact name (left shoulder, right elbow, etc.). When describing left and right body parts, always reference them from the person's perspective and not from the image. Light blue landmarks indicate the person's left side, and light yellow landmarks indicate the person's right side. Do not mix them up. This is regardless of image orientation. Use the trajectory in the image as your top reference, and use the cropped version for you to more clearly observe the trajectory itself. Do not make additional assumptions.\\
\indent 1a) Where is the blue square waypoint in the image? (this is where the motion starts) How close is the starting point in this segment to the nearest body waypoint, by looking at the textual information provided? If it is close, then the blue waypoint you see should be near that landmark in the image as well. Note that this is often the ending position of the last segment, unless this is the first segment, in which case it should be near the robot's gripper. Also note that this is different from the light blue body landmark. The landmarks are larger and have a paler color.\\
\indent 1b) Where is the red square waypoint in the image? (this is where the motion ends) How close is the ending point in this segment to the nearest body waypoint, by looking at the textual information provided? If it is close, then the red waypoint you see should be near that landmark in the image as well.\\
\indent 1c) What is the shape of the trajectory? Is it a straight line, or some other shape?\\
\indent 1d) Does the trajectory get close to some reference object in the environment, or pass through any human landmark? This can be either in the middle of the motion or at the end. If so, note the waypoint at which the robot gets closest to any landmarks/objects. Note that the trajectory may not be a straight line, so be sure to trace all the points from the start to the end, along the line segments.\\
\indent 1e) Is it moving towards something in the environment? If not, then is it moving in some nominal direction? You may use 'upward' and 'downward' to indicate +z and -z axes, as well as 'forward' and 'backward' to indicate the robot arm's reaching out and retracting back.\\
2) Consider the velocity of the robot end-effector. Recall that a brighter-green waypoint indicates higher speed, and darker-green point means lower speed. Length of line segments connecting two waypoints is also directly proportional to the speed. Base your response primarily on the textual velocity descriptions, and only use visual when you want to locate an interesting waypoint. Note that for this part, we define a "notable difference" or a "change" as anything greater than a two times difference. So if the speed decreases to more than half, it's called a slowdown, and if the speed increases to more than twice before, it's called a speedup.\\
\indent 2a) What is the speed of the robot near the start of this segment?\\
\indent 2b) What is the speed of the robot near the end of this segment?\\
\indent 2c) Is the starting speed notably different from the ending speed of the last segment?\\
\indent 2d) Is the ending speed in this segment notably different from the starting speed in this segment?\\
\indent 2e) If there is a notable change in velocity in this segment, either within the segment itself or as it transitions into this segment from the last, answer additionally the following:\\
\indent\indent 2e.i) At which waypoint does this change happen?\\
\indent\indent 2e.ii) Where is the robot end-effector position located within the environment, at this waypoint? Refer back to question 1 and the images to answer this.\\
\indent\indent 2e.iii) After this change, where is the robot moving towards?\\
3) Consider the force of the robot end effector. Recall that the line segments between two waypoints are colored according to the external forces as the end effector travels through that segment. In a gradient from blue to red, blue means lower force and red means higher force.\\
\indent 3a) Are there forces involved? If not, then you don't have to answer the rest of the questions in question group 3.\\
\indent 3b) If so, in what region is this force being applied on? Refer back to question 1 on the positions, as well as the images to answer this.\\
\indent 3c) Is that force changing throughout the segment? If so, how is it changing over the region that the force is applied on?\\
\indent 3d) The robot gripper is holding something in this segment; it is likely that the forces are being applied through this object being held, so what function could the held object serve, in relation to other components in the environment? What is robot planning to interact with?\\
\#\# Overall description for each segment:\\
After you answer these questions, associate the different components of this segment of motion---position, velocity, force---in the context of the environment. Generate a description of the robot's motion throughout this segment. When describing the motion direction, only use nominal directions when there isn't a clear object in the environment which the robot is moving to, or when the motion towards that object isn't a straight path. Otherwise, always describe the motion by referring to objects in the environment. If the motion is towards a body landmark, always use the exact name of the landmark, and don't refer to general body locations. The point is to be clear, specific, but also concise. Recall all the velocity and force changes that you noted. Where do these happen? Is it at a certain point or gradually? Include where the changes happen if there is a clear position. Keep your answer to this section under the heading ``\#\# Segment description''. Do not use this heading anywhere else in responding to previous numbered questions.
}
\end{tcolorbox}

\begin{tcolorbox}[title=Response ($\tau_1$), breakable,
    colback=response_frame!10, colframe=response_frame!50,
    left=2mm, right=2mm,       
    top=1mm, bottom=1mm]
{\setlength{\parindent}{1.5em}
\noindent 1) Position Analysis\\
\indent 1a) The blue square waypoint is near the robot's gripper, starting close to the left wrist.\\
\indent 1b) The red square waypoint is at the left wrist, indicating the end of the motion.\\
\indent 1c) The trajectory is a straight line.\\
\indent 1d) The trajectory ends at the left wrist.\\
\indent 1e) The motion is directed towards the left wrist.\\

\noindent 2) Velocity Analysis\\
\indent 2a) The speed near the start is 3.0 cm/s.\\
\indent 2b) The speed near the end is 2.0 cm/s.\\
\indent 2c) The starting speed is the same as the ending speed of the last segment, as this is the first segment.\\
\indent 2d) The ending speed is notably slower than the starting speed.\\
\indent 2e) Notable change in velocity:\\
\indent\indent 2e.i) The change happens at waypoint 7.\\
\indent\indent 2e.ii) At this waypoint, the robot is near the left wrist.\\
\indent\indent 2e.iii) After this change, the robot continues moving downward.\\
 
\noindent 3) Force Analysis\\
\indent 3a) There are no forces involved in this segment.\\
}

\#\# Segment Description\\
The robot moves in a straight line from its initial position towards the person's left wrist. The speed decreases as it approaches the wrist, indicating a careful approach. The trajectory ends at the left wrist without applying any force, suggesting the robot is positioning the held towel near the wrist, possibly for a handover or assistance task.
\end{tcolorbox}

\begin{tcolorbox}[title=Prompt ($\tau_2$), breakable,
    colback=prompt_frame!10, colframe=prompt_frame!50,
    left=2mm, right=2mm,       
    top=1mm, bottom=1mm]
\# \textbf{Understanding your planned trajectory}\\
\texttt{[...]}\\

\# \textbf{Kinematic description for image 2}\\
At the start of image {\color{prompt_frame!125}2}, the gripper is {\color{prompt_frame!125}closed}.\\
\#\# Position descriptions\\
At waypoint {\color{prompt_frame!125}9}, the position of the gripper center is at {\color{prompt_frame!125}(-16.0, -49.2, 74.2)}. From waypoint {\color{prompt_frame!125}9} to waypoint {\color{prompt_frame!125}10}, the general motion of the robot is {\color{prompt_frame!125}(0.6, 1.1, 1.6)} centimeters. \texttt{[...]} At waypoint {\color{prompt_frame!125}39}, the position of the gripper center is at {\color{prompt_frame!125}(-39.3, -52.8, 84.8)}. \\
\#\# Velocity descriptions\\
From waypoint {\color{prompt_frame!125}9} to waypoint {\color{prompt_frame!125}10}, the average velocity of the robot is {\color{prompt_frame!125}1.0} centimeters per second. \texttt{[...]} From waypoint {\color{prompt_frame!125}38} to waypoint {\color{prompt_frame!125}39}, the average velocity of the robot is {\color{prompt_frame!125}1.0} centimeters per second. \\
\#\# Force descriptions\\
From waypoint {\color{prompt_frame!125}9} to waypoint {\color{prompt_frame!125}10}, the force magnitude of the robot end-effector is: {\color{prompt_frame!125}[1.00, 1.00, 1.01, 1.01, 1.01, 1.02, 1.02, \texttt{...}, 1.14, 1.14, 1.15, 1.15]}. \texttt{[...]} From waypoint {\color{prompt_frame!125}38} to waypoint {\color{prompt_frame!125}39}, the force magnitude of the robot end-effector is: {\color{prompt_frame!125}[3.09, 3.09, 3.09, 3.10, 3.10, 3.10, 3.11, \texttt{...}, 3.15, 3.16, 3.16, 3.16]}. \\
\#\# Human body landmark positions\\
The following are the positions for each of the detected body landmarks in the image: {\color{prompt_frame!125}left wrist: (-16.0, -49.2, 74.2) centimeters, \texttt{[...]}}\\
The first waypoint in this segment is closest to the {\color{prompt_frame!125}left wrist} at a distance of {\color{prompt_frame!125}0.0} centimeters. The last waypoint in this segment is closest to the {\color{prompt_frame!125}left elbow} at a distance of {\color{prompt_frame!125}1.0} centimeters. \\

\# \textbf{Task descriptions}\\
\texttt{[...]}
\end{tcolorbox}

\begin{tcolorbox}[title=Response ($\tau_2$), breakable,
    colback=response_frame!10, colframe=response_frame!50,
    left=2mm, right=2mm,       
    top=1mm, bottom=1mm]
{\setlength{\parindent}{1.5em}
\noindent 1) Position Analysis\\
\indent 1a) The blue square waypoint starts at the left wrist.\\
\indent 1b) The red square waypoint ends near the left elbow.\\
\indent 1c) The trajectory is a straight line.\\
\indent 1d) The trajectory passes close to the left elbow.\\
\indent 1e) The motion is directed towards the left elbow.\\

\noindent 2) Velocity Analysis\\
\indent 2a) The speed near the start is 1.0 cm/s.\\
\indent 2b) The speed near the end is 1.0 cm/s.\\
\indent 2c) The starting speed is slower than the ending speed of the last segment.\\
\indent 2d) The ending speed is the same as the starting speed in this segment.\\
\indent 2e) There is no notable change in velocity within this segment.\\
 
\noindent 3) Force Analysis\\
\indent 3a) Yes, there are forces involved.\\
\indent 3b) The force is applied throughout the motion from the left wrist to the left elbow.\\
\indent 3c) The force gradually increases throughout the segment.\\
\indent 3d) The held towel might be used to apply gentle pressure or cleaning action along the arm.\\
}

\#\# Segment Description\\
The robot moves in a straight line from the left wrist towards the left elbow. The speed remains constant throughout the motion. A gradually increasing force is applied, suggesting the robot is using the towel to gently interact with the arm, possibly for cleaning or applying pressure.
\end{tcolorbox}

For $\tau_3$, note the additional question at the bottom where CoRI asks about the overall intention and any user cooperation needed.

\begin{tcolorbox}[title=Prompt ($\tau_3$), breakable,
    colback=prompt_frame!10, colframe=prompt_frame!50,
    left=2mm, right=2mm,       
    top=1mm, bottom=1mm]
\# \textbf{Understanding your planned trajectory}\\
\texttt{[...]}\\

\# \textbf{Kinematic description for image 3}\\
At the start of image {\color{prompt_frame!125}3}, the gripper is closed.\\
\#\# Position descriptions\\
At waypoint {\color{prompt_frame!125}39}, the position of the gripper center is at {\color{prompt_frame!125}(-39.3, -52.8, 84.8)}. From waypoint {\color{prompt_frame!125}39} to waypoint {\color{prompt_frame!125}40}, the general motion of the robot is {\color{prompt_frame!125}(0.0, 0.1, 1.8)} centimeters. \texttt{[...]} At waypoint {\color{prompt_frame!125}63}, the position of the gripper center is at {\color{prompt_frame!125}(-4.3, -39.1, 79.3)}. \\
\#\# Velocity descriptions\\
From waypoint {\color{prompt_frame!125}39} to waypoint {\color{prompt_frame!125}40}, the average velocity of the robot is {\color{prompt_frame!125}2.0} centimeters per second. \texttt{[...]} From waypoint {\color{prompt_frame!125}62} to waypoint {\color{prompt_frame!125}63}, the average velocity of the robot is {\color{prompt_frame!125}4.0} centimeters per second. \\
\#\# Force descriptions\\
{\color{prompt_frame!125}There is no external force planned during this section.} \\
\#\# Human body landmark positions\\
The following are the positions for each of the detected body landmarks in the image: {\color{prompt_frame!125}left wrist: (-16.0, -49.2, 74.2) centimeters, \texttt{[...]}}\\
The first waypoint in this segment is closest to the {\color{prompt_frame!125}left elbow} at a distance of {\color{prompt_frame!125}1.0} centimeters. The last waypoint in this segment is closest to the {\color{prompt_frame!125}left wrist} at a distance of {\color{prompt_frame!125}16.3} centimeters. \\

\# \textbf{Task descriptions}\\
\texttt{[...]}\\
\#\# Overall intention for entire trajectory:\\
What is the overall intention of the robot, now that you have seen all the planned trajectories? This should be closely related to what you identified that the robot is holding between the grippers. Reason about its functionality within the trajectory. Focus on the intention and not the motion. Keep your answer to this question under the heading ``\#\# Overall intention''.\\
Additionally, what does the user need to do in each segment that will allow this overall intention to be achieved? It is fine if there is no cooperation needed during some/all segments. Keep your answer to this question under the heading ``\#\# User cooperation''.
\end{tcolorbox}

\begin{tcolorbox}[title=Response ($\tau_3$), breakable,
    colback=response_frame!10, colframe=response_frame!50,
    left=2mm, right=2mm,       
    top=1mm, bottom=1mm]
{\setlength{\parindent}{1.5em}
\noindent 1) Position Analysis\\
\indent 1a) The blue square waypoint starts near the left elbow.\\
\indent 1b) The red square waypoint ends near the left wrist.\\
\indent 1c) The trajectory is a straight line.\\
\indent 1d) The trajectory gets close to the left elbow and ends near the left wrist.\\
\indent 1e) The motion is directed downward towards the left wrist.\\

\noindent 2) Velocity Analysis\\
\indent 2a) The speed near the start is 2.0 cm/s.\\
\indent 2b) The speed near the end is 4.0 cm/s.\\
\indent 2c) The starting speed is slower than the ending speed of the last segment.\\
\indent 2d) The ending speed is notably faster than the starting speed in this segment.\\
\indent 2e) Notable change in velocity:\\
\indent\indent 2e.i) The change happens at waypoint 45.\\
\indent\indent 2e.ii) At this waypoint, the robot is moving downward from the left elbow.\\
\indent\indent 2e.iii) After this change, the robot continues moving downward towards the left wrist.\\
 
\noindent 3) Force Analysis\\
\indent 3a) There are no forces involved in this segment.\\
}

\#\# Segment Description\\
The robot moves in a straight line from the left elbow towards the left wrist. The speed increases as it moves downward, indicating a swift motion. No force is applied during this segment, suggesting the robot is repositioning the towel or returning to a neutral position.\\

\#\# Overall Intention\\
The overall intention of the robot is to use the towel to interact with the person's left arm, possibly for cleaning or applying gentle pressure. The robot carefully approaches the arm, applies a gentle force along the arm, and then repositions itself.
\end{tcolorbox}

\subsection{Communication Generation Stage}

In this stage, CoRI extracts the following text from previous VLM responses: 1) all the response to the environment comprehension stage, 2) text under ``\#\# Segment Description'' headings, and 3) text under the ``\#\# Overall Intention'' heading. A prompt is then built around these text, and a reasoning LLM is queried to generate the final communication. Parts of the prompt that is extracted from previous stages are colored in {\color{prompt_frame!125}navy}.

\begin{tcolorbox}[title=Prompt, breakable,
    colback=prompt_frame!10, colframe=prompt_frame!50,
    left=2mm, right=2mm,       
    top=1mm, bottom=1mm]
You are a robot, and here is what you currently observe in the environment:\\
{\color{prompt_frame!125}
\#\#\# Environment:\\
- A person is sitting on a bed or couch, wearing a maroon hoodie and jeans.\\
- The person's body joints and facial landmarks are labeled with blue and yellow markers, indicating left and right sides.\\
- The setting appears to be indoors, possibly a room or living area.\\
\#\# Object in Gripper:\\
- The robot is holding a white cloth or towel between its grippers.\\
\#\#\# Possible Function:\\
\color{prompt_frame!125}- If the robot does not let go of the towel, it could be used for cleaning or wiping purposes, such as assisting the person with personal hygiene or tidying up the surrounding area.\\
}

You also have a trajectory planned to interact with the environment. It is broken down into segments, and here are the descriptions for each segment:\\
{\color{prompt_frame!125}
- Segment 1: The robot moves in a straight line from its initial position towards the person's left wrist. The speed decreases as it approaches the wrist, indicating a careful approach. The trajectory ends at the left wrist without applying any force, suggesting the robot is positioning the held towel near the wrist, possibly for a handover or assistance task.\\
- Segment 2: The robot moves in a straight line from the left wrist towards the left elbow. The speed remains constant throughout the motion. A gradually increasing force is applied, suggesting the robot is using the towel to gently interact with the arm, possibly for cleaning or applying pressure.\\
- Segment 3: The robot moves in a straight line from the left elbow towards the left wrist. The speed increases as it moves downward, indicating a swift motion. No force is applied during this segment, suggesting the robot is repositioning the towel or returning to a neutral position.\\
}

Additionally, from the motion and the observation, the intention of the entire trajectory is tentatively deduced to be: {\color{prompt_frame!125}``The overall intention of the robot is to use the towel to interact with the person's left arm, possibly for cleaning or applying gentle pressure. The robot carefully approaches the arm, applies a gentle force along the arm, and then repositions itself.''}\\

Further, the user cooperation in each segment needed in order to achieve the overall intention is tentatively deduced to be:\\
{\color{prompt_frame!125}
- Segment 1: The user should keep their left arm steady to allow the robot to approach the wrist accurately.\\
- Segment 2: The user should maintain their arm position to let the robot apply the towel along the arm.\\
- Segment 3: No cooperation is needed as the robot repositions itself.\\
}

I now need you to do the following: first rephrase the overall intention and user cooperation so that they are very friendly and suitable in an assistive setting. Don't use emotionless words like ``task''. Also ensure that these make sense within the context of each segment's description, and resolve any inconsistencies. Take the most likely intention and don't give ambiguous answers.\\
Next, I need you to output one concise sentence per segment, as a statement that speaks to the user before you execute the motion for that segment. Include the overall \textbf{intention} where appropriate, and mention any active human cooperation required. Pay attention to when the robot actually plans to engage in contact, and make sure your statements are consistent with this--when the description for a segment says no force, it directly means that there is no contact, either the contact has not started yet, or the contact is already finished. Note that with the exception of the first segment, the statement will be said *while* the robot is moving, so it'll be the best approach to include the high-level intention in the first statement and omit it in the remaining statements. In terms of the motion, each statement should only talk about the motion within that segment. In summary, each of your sentences should contain the following information (in no particular order):\\
- where the robot is moving to within the segment (starting position can in general be omitted, so no need for phrases like ``from $<$some place$>$'')\\
- \emph{If the shape of the trajectory in the segment is more complicated than a straight line}, then mention it, otherwise just say where you'll move to without saying ``in a straight line''\\
- \emph{If velocity is not near-constant}, then changes and trends in velocity, and where they happen. If the description says the velocity/speed is constant/steady within the segment, don't mention anything about the speed all together.\\
- \emph{If there is any force}, then changes and trends in force, and where they happen. If the description says no force/pressure, it simply means that no contact is planned for the robot to actively engage with the environment, and in this case don't mention anything about force/contact all together.\\
- \emph{If needed}, any behavior the human should do in this segment to achieve the overall intention. Be specific for this one--you need to say \textbf{exactly} what the human should do to cooperate. If necessary, modify the previously determined user cooperation.\\
- If appropriate, the intention of the segment, but be simple and basic\\
Try to weave these different information together using simple language, and use qualitative descriptions instead of numerical values. Be concise and direct. For example, simply stating ``getting faster'' is an implication that you are starting slow, so no need to elaborate by saying things like ``starting slow and then speeding up''. If the description says something like ``no force'' or ``no contact'', it simply means that the robot is not planning to touch anything, so if the previous segment contains force and the current segment does not, it simply means that the contact is finished. Do not get this confused. You can ignore or condense extra information from the provided descriptions if they don't fit in one of the categories listed above. Suppose you are speaking to a ten-year-old. Simplify as much as you can. Speak in a friendly and helpful manner. Format your output as ``Statement x: $<$your sentence$>$''.
\end{tcolorbox}

\begin{tcolorbox}[title=Response, breakable,
    colback=response_frame!10, colframe=response_frame!50,
    left=2mm, right=2mm,       
    top=1mm, bottom=1mm]
Overall Intention and Cooperation:\\
My aim is to help you feel comfortable by gently using the soft towel on your left arm—whether that means a light clean or a soothing touch. To work well together, please keep your left arm steady as I approach and help along your arm; once I finish the gentle touch, you can simply relax.\\

Statement 1: I'm moving directly toward your left wrist and slowing down as I get closer so I can carefully position the towel for a gentle clean—please keep your left arm still as I approach.\\
Statement 2: I'm now moving from your left wrist to your left elbow at a steady pace, gradually increasing my touch with the towel for a gentle sweep along your arm—please continue to hold your arm steady.\\
Statement 3: I'm quickly moving from your left elbow back to your left wrist without any touch, simply repositioning the towel—now you can relax.
\end{tcolorbox}

Note that we only extract the sentence after each ``Statement $k$'' as the statement to narrate to the user.

\section{Trajectory Plans, Ground Truths, and Sample Outputs}

For each of the three tasks evaluated in the user study (simulated bathing, simulated shaving, and feeding), we design two different trajectories. In this section, for each of the trajectories, we provide: 1) an example trajectory, shown in the same overlay fashion as what the VLM reasons on, 2) the ground-truth paragraph for each segment in the trajectory, 3) the one-sentence summary for each paragraph, generated by a reasoning LLM (o3-mini), and 4) sample communications generated by CoRI, for different participants throughout our study.

\subsection{Bathing}

\textbf{Trajectory 1}

\begin{figure}[tbh]
    \centering
    \begin{subfigure}[t]{0.28\textwidth}  
      \centering
      \includegraphics[width=\linewidth]{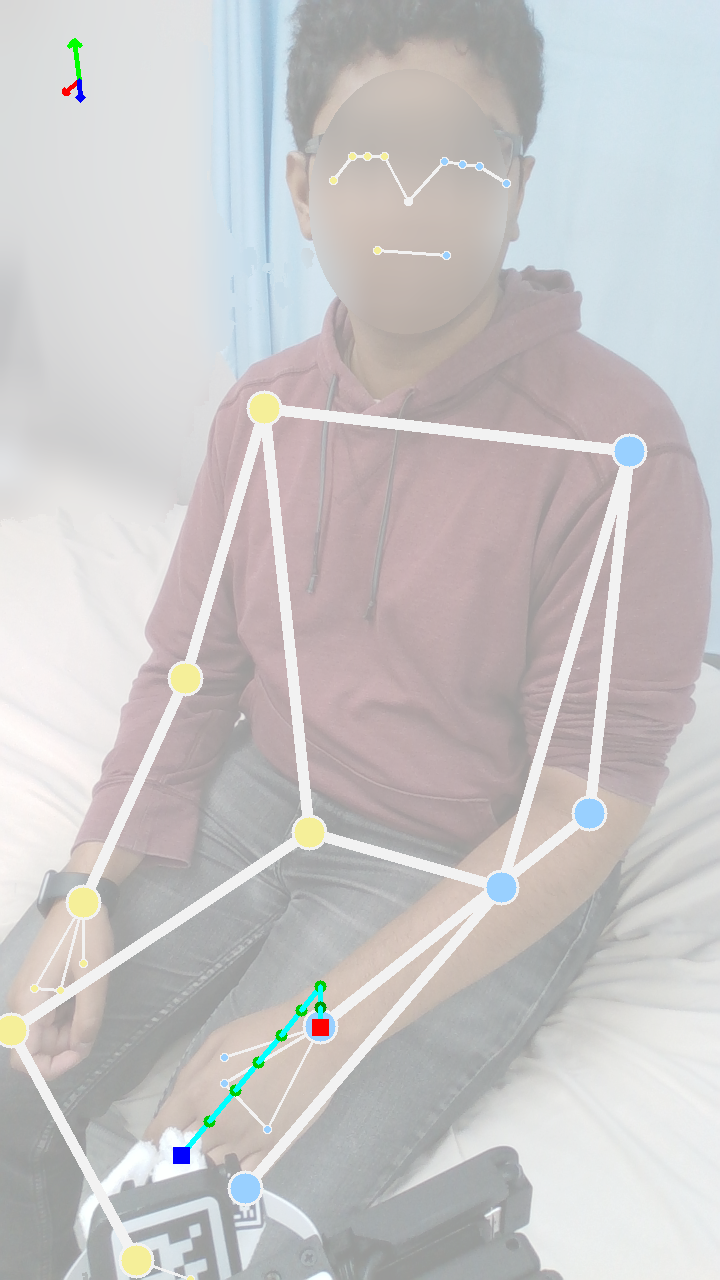}  
      \caption{Segment 1}
    \end{subfigure}\hfill
    \begin{subfigure}[t]{0.28\textwidth}
      \centering
      \includegraphics[width=\linewidth]{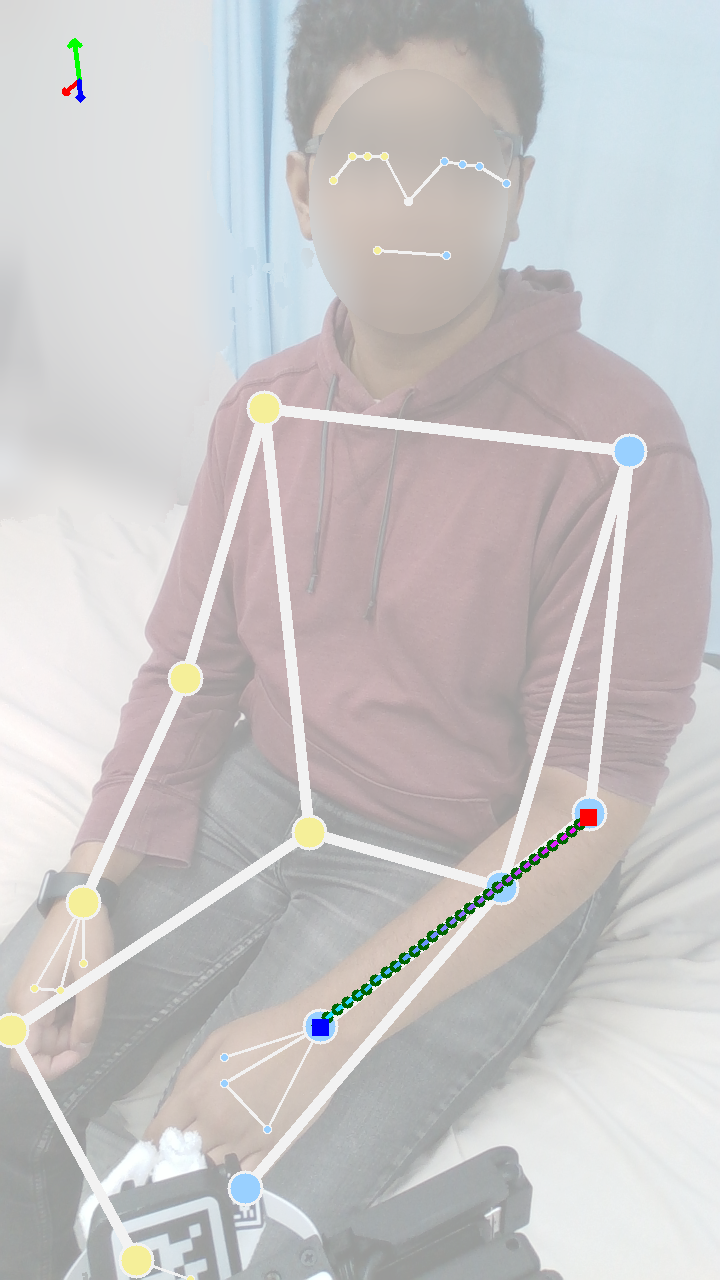}
      \caption{Segment 2}
    \end{subfigure}\hfill
    \begin{subfigure}[t]{0.28\textwidth}
      \centering
      \includegraphics[width=\linewidth]{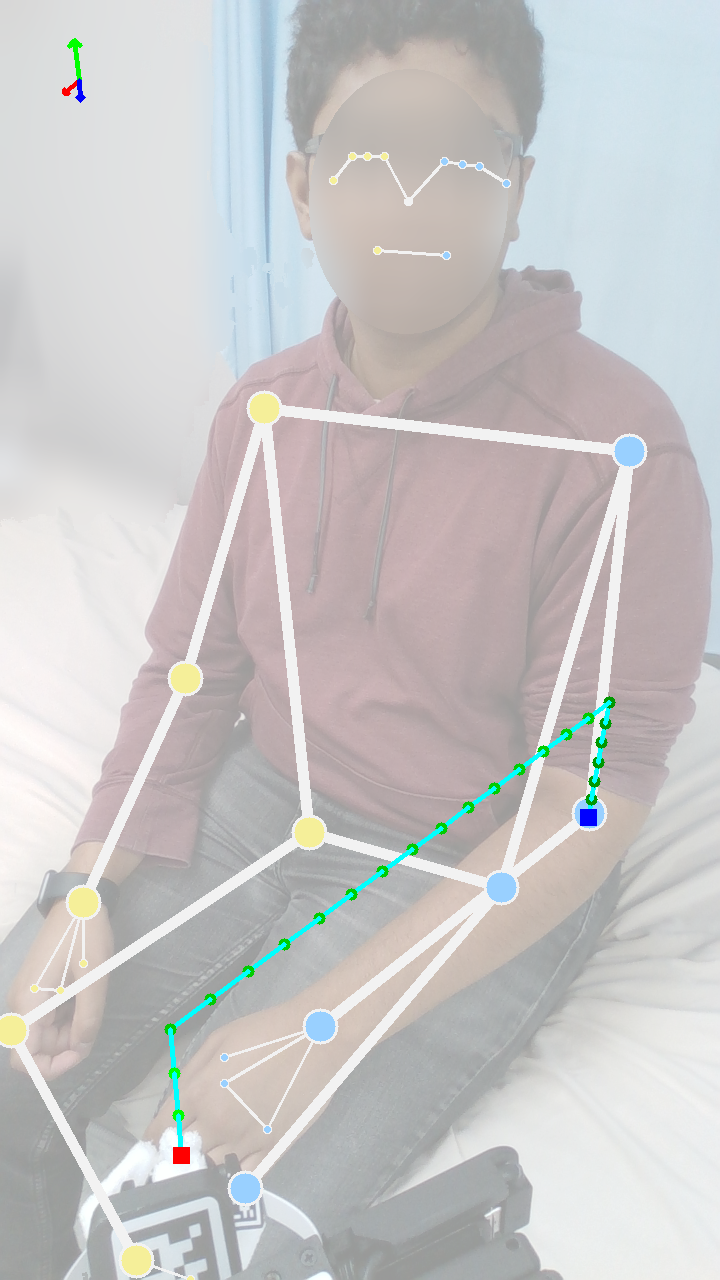}
      \caption{Segment 3}
    \end{subfigure}
    \caption{Example overlay visualization of trajectory 1 in the bathing task, for participant 3.}
    \label{fig:bathing-1}
\end{figure}

\begin{tcolorbox}[title=Ground Truth Paragraphs, breakable,
    colback=ground_truth_frame!10, colframe=ground_truth_frame!50,
    left=2mm, right=2mm,       
    top=1mm, bottom=1mm]
Statement for segment 1: I'm slowly moving the soft towel that I'm holding from my starting position towards your left wrist, starting in a straight, gentle diagonal path to a spot just above your wrist, and then moving straight down slightly near the end as I'm near you, to gently place/rest the towel right at your left wrist. I'll start at a moderate speed and slow down as I get close. I'm not touching you for the most part, but you may feel a light, gentle, caring contact near the end. I'm just carefully positioning myself and the towel so that everything is ready before beginning our gentle cleaning/freshening/caring process of a soothing wipe. Please keep your left wrist and left arm still, open, and visible, so that I can reach exactly where I need in order to help you, and you can just sit back and relax as I do this approach on my own.\\

Statement for segment 2: I'm now slowly and carefully gliding the towel up along your left arm, in a straight line from your left wrist to your left elbow, maintaining a constant low speed, and gradually increasing the pressure on my gentle touch as I move, so that I can ensure proper cleaning through the wipe. Please continue to hold your left arm still and open, and just stay comfortable, as I work to complete the gentle cleaning, caring process along your forearm.\\

Statement for segment 3: Lastly, I'm now lifting the towel away from your left elbow in a curved path, first moving slowly and slightly upward near your left elbow along your upper arm, and then smoothly and quickly retracting my hand with the cloth away from your body, returning to my base starting position above your lap in a straight diagonal, moving faster as I leave, without any pressure. I'm not touching you anymore, so there's nothing you need to do--you can simply relax in comfort now as I finish the cleaning, but keep your arm steadily as it is as I get ready for my next wipe.
\end{tcolorbox}

\begin{tcolorbox}[title=Summaries from LLM, breakable,
    colback=summary_frame!10, colframe=summary_frame!50,
    left=2mm, right=2mm,       
    top=1mm, bottom=1mm]
Statement for segment 1: I'll move the towel along a gentle diagonal then straight-down path to rest it on your left wrist, slowing from moderate to near stop with only light contact at the end; please keep your left arm still, open, and visible.\\

Statement for segment 2: I'll glide the towel straight up your left forearm to your elbow, gradually increasing pressure for effective cleaning; please keep your left arm still and open.\\

Statement for segment 3: I'll lift the towel away from your elbow in a slow, curved upward path, then accelerate back along a diagonal to my lap; please keep your arm steady.
\end{tcolorbox}

\begin{tcolorbox}[title=Sample communication generated from CoRI (\textbf{P3}), breakable,
    colback=comm_frame!10, colframe=comm_frame!50,
    left=2mm, right=2mm,       
    top=1mm, bottom=1mm]
Statement for segment 1: I'm moving directly toward your left wrist and slowing down as I get closer so I can carefully position the towel for a gentle clean—please keep your left arm still as I approach.\\

Statement for segment 2: I'm now moving from your left wrist to your left elbow at a steady pace, gradually increasing my touch with the towel for a gentle sweep along your arm—please continue to hold your arm steady.\\

Statement for segment 3: I'm quickly moving from your left elbow back to your left wrist without any touch, simply repositioning the towel—now you can relax.
\end{tcolorbox}

\begin{tcolorbox}[title=Sample communication generated from CoRI (\textbf{P9}), breakable,
    colback=comm_frame!10, colframe=comm_frame!50,
    left=2mm, right=2mm,       
    top=1mm, bottom=1mm]
Statement for segment 1: I'm carefully moving my towel in a mostly straight path, then lowering it softly to your left wrist so I can start the gentle cleaning—please keep your wrist still.\\

Statement for segment 2: I'm now gently moving the towel upward along your forearm toward your left elbow while slowly increasing my pressure to wipe it nicely—please keep your arm extended and steady.\\

Statement for segment 3: I'm briefly lifting near your elbow and then quickly moving my towel away in a diagonal and downward path to finish the cleaning, so you can simply relax.
\end{tcolorbox}

\begin{tcolorbox}[title=Sample communication generated from CoRI (\textbf{P14}), breakable,
    colback=comm_frame!10, colframe=comm_frame!50,
    left=2mm, right=2mm,       
    top=1mm, bottom=1mm]
Statement for segment 1: I'm going to gently move the soft towel straight down onto your left wrist, slowing down as I near it to begin caring for your arm—please just relax your wrist.\\

Statement for segment 2: Now, I'm slowly sliding the towel along your forearm toward your elbow, gently increasing the pressure as I follow your arm's contour; please keep your left arm steady.\\

Statement for segment 3: Finally, I'll lift and curve the towel away from your arm, speeding up as I move it safely above your lap, so you can simply remain relaxed while I finish.
\end{tcolorbox}

\textbf{Trajectory 2}

\begin{figure}[bh]
    \centering
    \begin{subfigure}[t]{0.28\textwidth}  
      \centering
      \includegraphics[width=\linewidth]{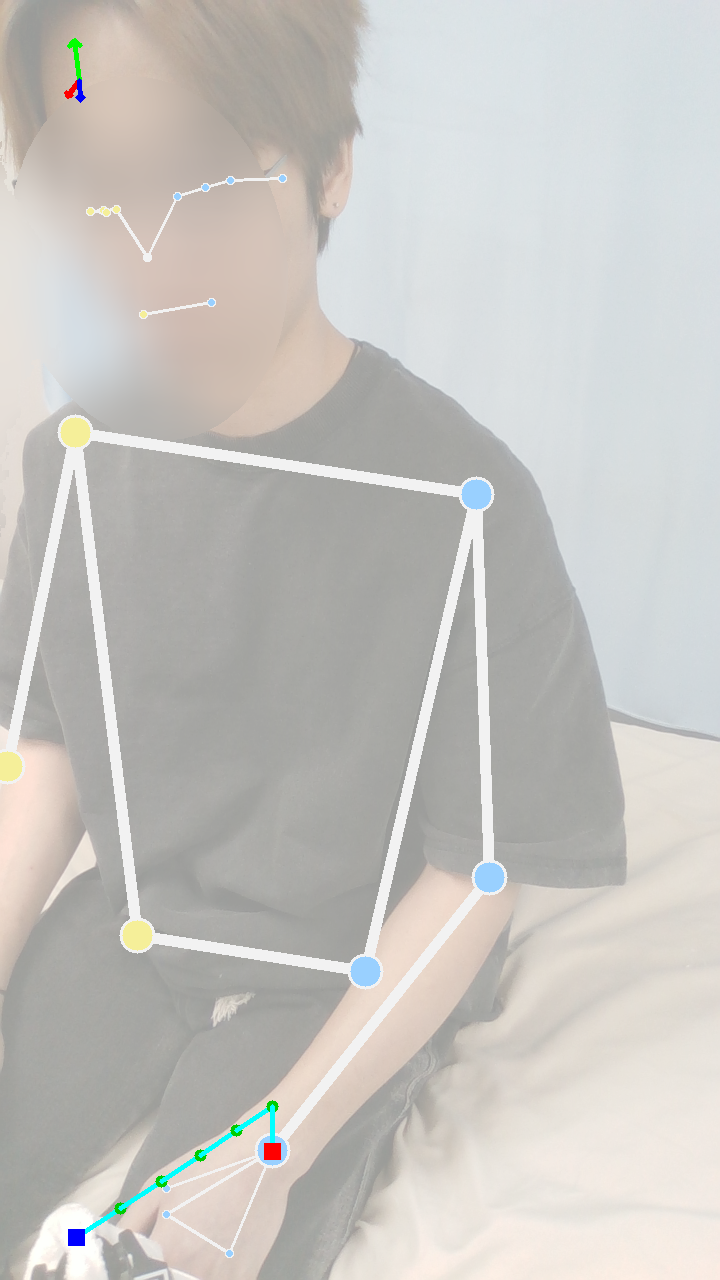}  
      \caption{Segment 1}
    \end{subfigure}\hfill
    \begin{subfigure}[t]{0.28\textwidth}
      \centering
      \includegraphics[width=\linewidth]{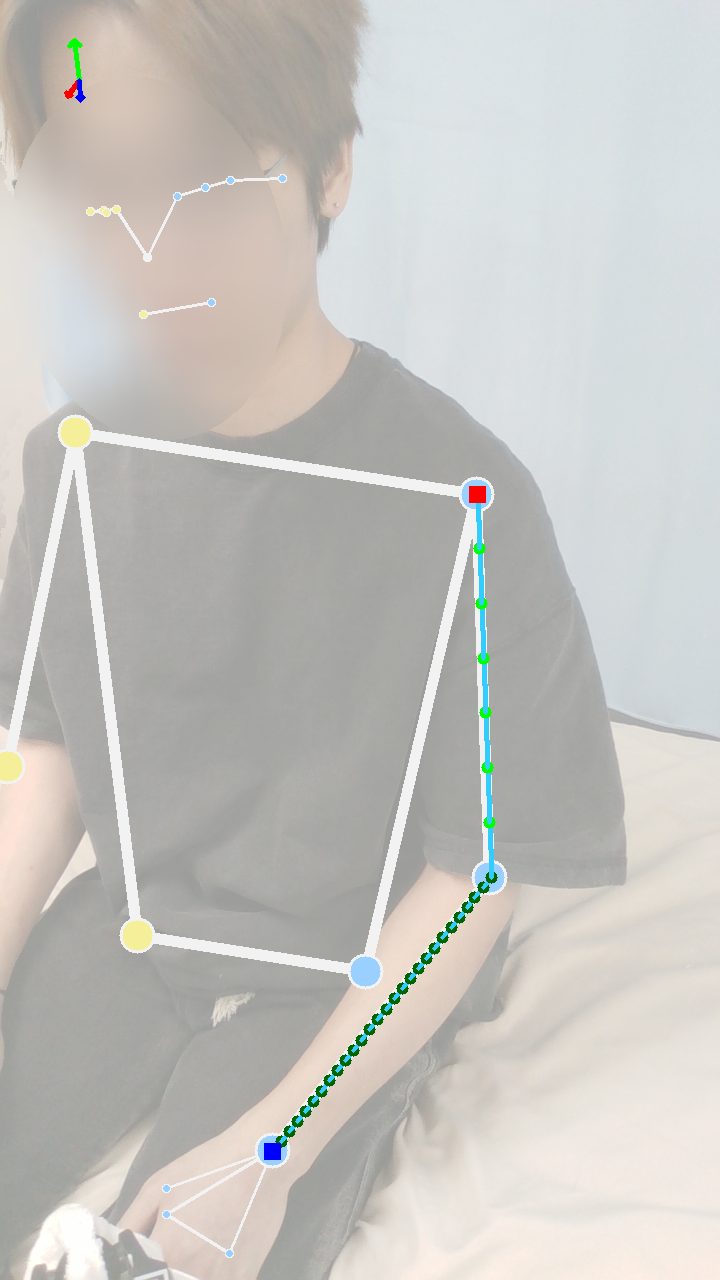}
      \caption{Segment 2}
    \end{subfigure}\hfill
    \begin{subfigure}[t]{0.28\textwidth}
      \centering
      \includegraphics[width=\linewidth]{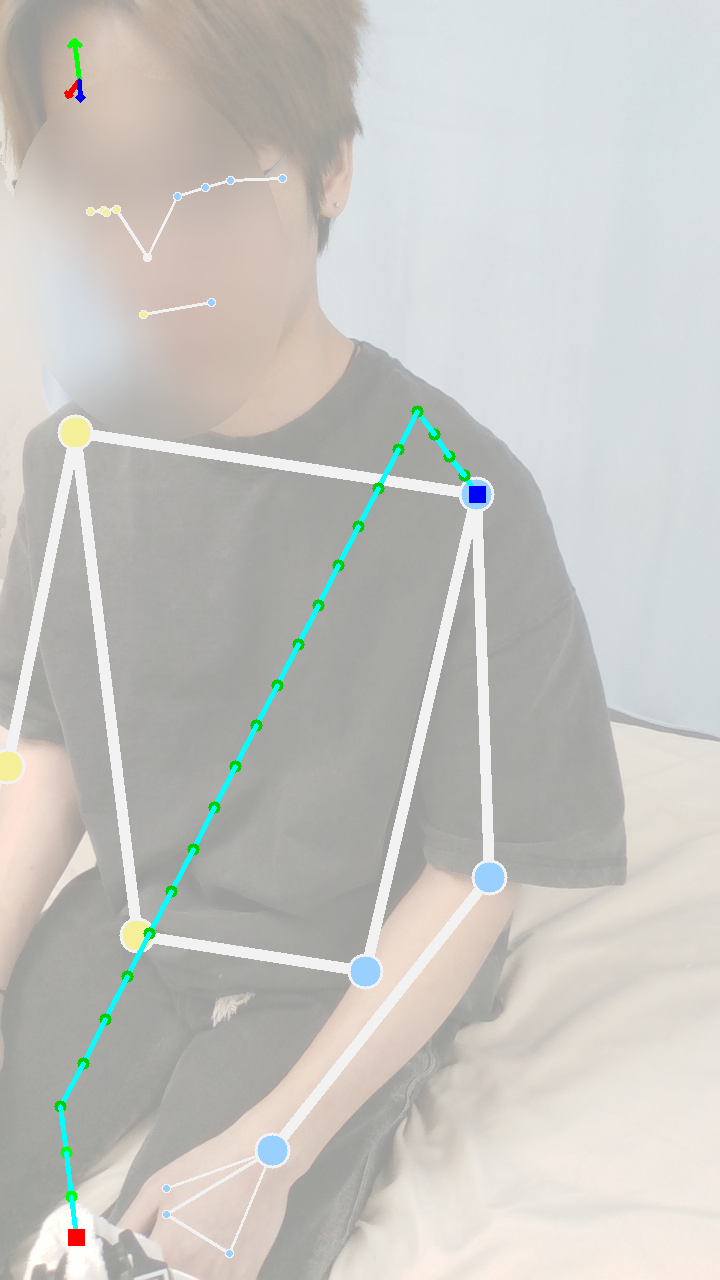}
      \caption{Segment 3}
    \end{subfigure}
    \caption{Example overlay visualization of trajectory 2 in the bathing task, for participant 6.}
    \label{fig:bathing-2}
\end{figure}

\begin{tcolorbox}[title=Ground Truth Paragraphs, breakable,
    colback=ground_truth_frame!10, colframe=ground_truth_frame!50,
    left=2mm, right=2mm,       
    top=1mm, bottom=1mm]
Statement for segment 1: I'm slowly moving the soft towel that I'm holding (or some other soft, fluffy item) from my starting position towards your left wrist, starting in a straight, gentle diagonal line to a spot just above your wrist, and then moving straight down slightly near the end as I'm near you, to gently place/rest the towel right at your left wrist. I'll start at a moderate speed and slow down as I get close. I'm not touching you or applying pressure yet for the most part, but you may feel a light, gentle, caring contact near the end. I'm just carefully positioning myself and the towel so that everything is ready before beginning our gentle cleaning/freshening/caring process of a soothing wipe. Please keep your left wrist and left arm still, open, and visible, so that I can reach exactly where I need in order to help you, and you can just sit back and relax as I do this approach on my own.\\

Statement for segment 2: I'm now slowly and carefully gliding the towel up along your left arm, from your left wrist to your left elbow in a straight line and then to your left shoulder, following the natural contour of your arm in a curve. I'll start slow and move faster near your elbow as I get close to your shoulder in the end, while softly applying a constant, gentle pressure with my touch. Please continue to hold your left arm steady and open, and just stay comfortable, as I work to safely complete the gentle cleaning, caring process of wiping along your arm.\\

Statement for segment 3: Lastly, I'm now lifting the towel away from your left shoulder in a curved path, first moving slowly and steadily and then smoothly and quickly retracting the cloth away from your body, moving down in a straight diagonal path to return to my base starting position above your lap around your wrist, moving faster as I leave, without any pressure. I'm not touching or pressing you anymore, so there's nothing you need to do--you can simply relax in comfort now as I finish the cleaning, but keep your arm steadily as it is as I get ready for my next wipe.
\end{tcolorbox}

\begin{tcolorbox}[title=Summaries from LLM, breakable,
    colback=summary_frame!10, colframe=summary_frame!50,
    left=2mm, right=2mm,       
    top=1mm, bottom=1mm]
Statement for segment 1: I will move the towel in a gentle diagonal toward and then slightly down to rest at your left wrist, slowing as I approach—please keep your left wrist and arm still, open, and visible so I can position the towel before cleaning.\\

Statement for segment 2: I will glide the towel up along the natural curve of your left arm to your shoulder, starting slowly and speeding up near your elbow while applying constant gentle pressure—please continue to hold your arm steady and open, and stay comfortable as I wipe.\\

Statement for segment 3: I will lift the towel away from your shoulder in a curved path and retract it diagonally back to my rest position, accelerating as I leave with no contact—please keep your arm steady and relax.
\end{tcolorbox}

\begin{tcolorbox}[title=Sample communication generated from CoRI (\textbf{P6}), breakable,
    colback=comm_frame!10, colframe=comm_frame!50,
    left=2mm, right=2mm,       
    top=1mm, bottom=1mm]
Statement for segment 1: I'm moving straight to your left wrist with the towel to get ready for a gentle cleaning—please just relax as I position myself.\\

Statement for segment 2: Now I'm softly running the towel along your left arm from your wrist to your shoulder, following its natural curve and moving a bit faster after your elbow; please keep your left arm still so I can do this smoothly.\\

Statement for segment 3: I'm now moving the towel away from your shoulder along a smooth, curved path, finishing the cleaning without any contact—feel free to relax.
\end{tcolorbox}

\begin{tcolorbox}[title=Sample communication generated from CoRI (\textbf{P10}), breakable,
    colback=comm_frame!10, colframe=comm_frame!50,
    left=2mm, right=2mm,       
    top=1mm, bottom=1mm]
Statement for segment 1: I'm moving straight over to your left wrist with my towel to begin a gentle clean; please keep your left arm steady.\\

Statement for segment 2: I'm now gliding the towel along your arm, starting slowly at your wrist and picking up a bit of speed after your elbow as I head to your left shoulder; please hold your arm still.\\

Statement for segment 3: I'm gently retracting the towel along your arm, moving past your wrist with a little extra speed as I finish up, so please continue to keep your arm calm.
\end{tcolorbox}

\begin{tcolorbox}[title=Sample communication generated from CoRI (\textbf{P15}), breakable,
    colback=comm_frame!10, colframe=comm_frame!50,
    left=2mm, right=2mm,       
    top=1mm, bottom=1mm]
Statement for segment 1: I'm moving the towel in a smooth, straight line to softly reach your left wrist—please keep your arm still so I can begin our gentle comfort.\\

Statement for segment 2: Now, I'm gliding the towel up your arm from your wrist to your shoulder with a gentle, steady pressure that picks up a bit after your elbow; please continue to keep your arm relaxed.\\

Statement for segment 3: I'm now moving the towel diagonally from your shoulder back toward your wrist without pressing, as I return to a neutral position—just relax and enjoy the movement.
\end{tcolorbox}

\subsection{Shaving}

\textbf{Trajectory 1}

\begin{figure}[bh]
    \centering
    \begin{subfigure}[t]{0.28\textwidth}  
      \centering
      \includegraphics[width=\linewidth]{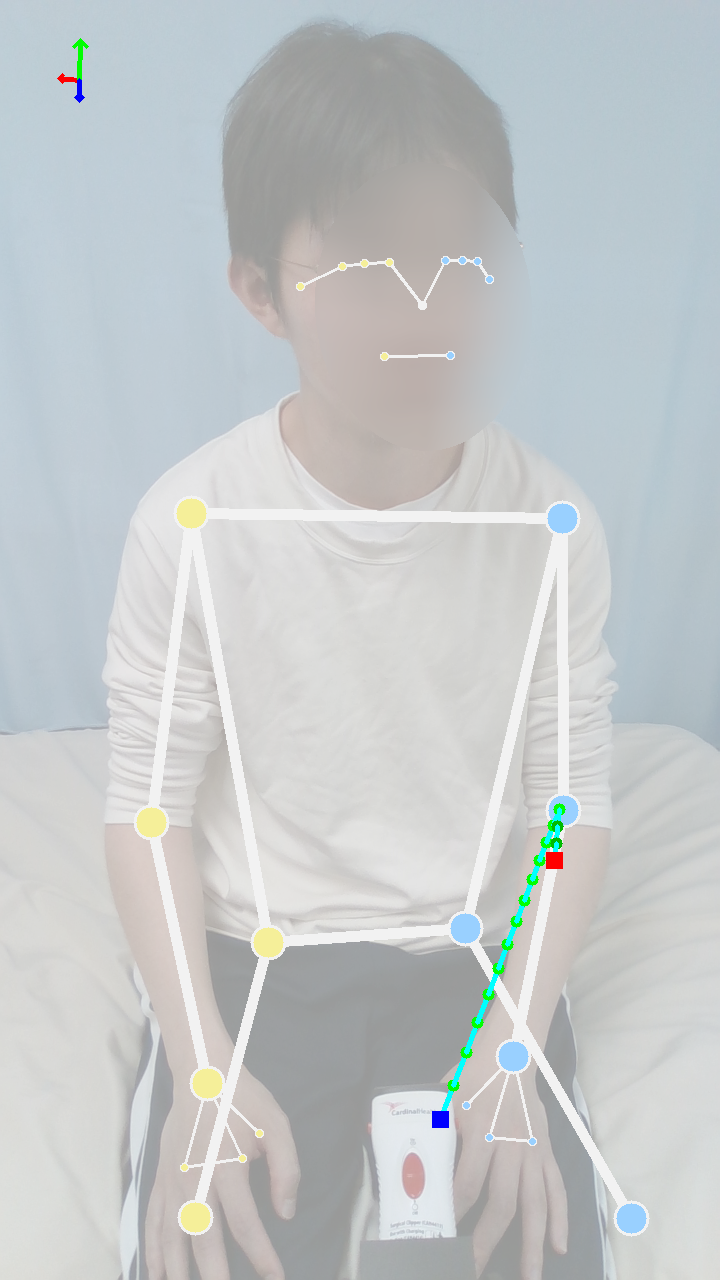}  
      \caption{Segment 1}
    \end{subfigure}\hfill
    \begin{subfigure}[t]{0.28\textwidth}
      \centering
      \includegraphics[width=\linewidth]{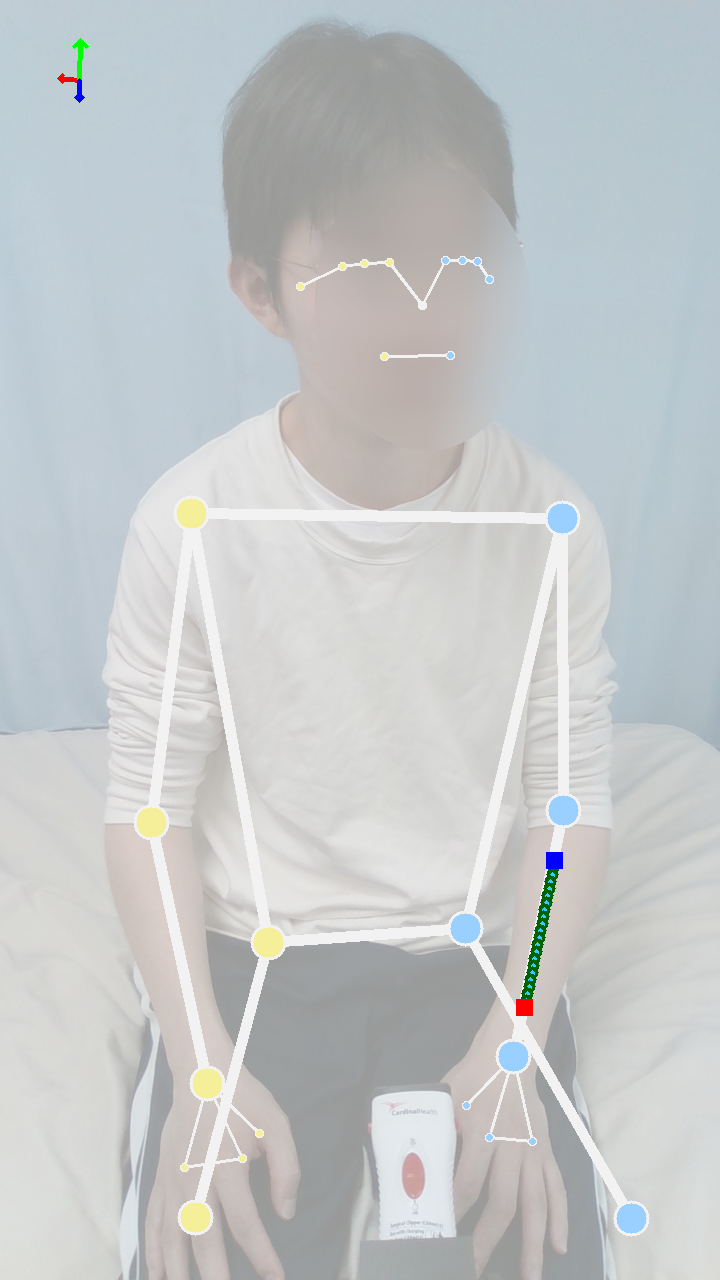}
      \caption{Segment 2}
    \end{subfigure}\hfill
    \begin{subfigure}[t]{0.28\textwidth}
      \centering
      \includegraphics[width=\linewidth]{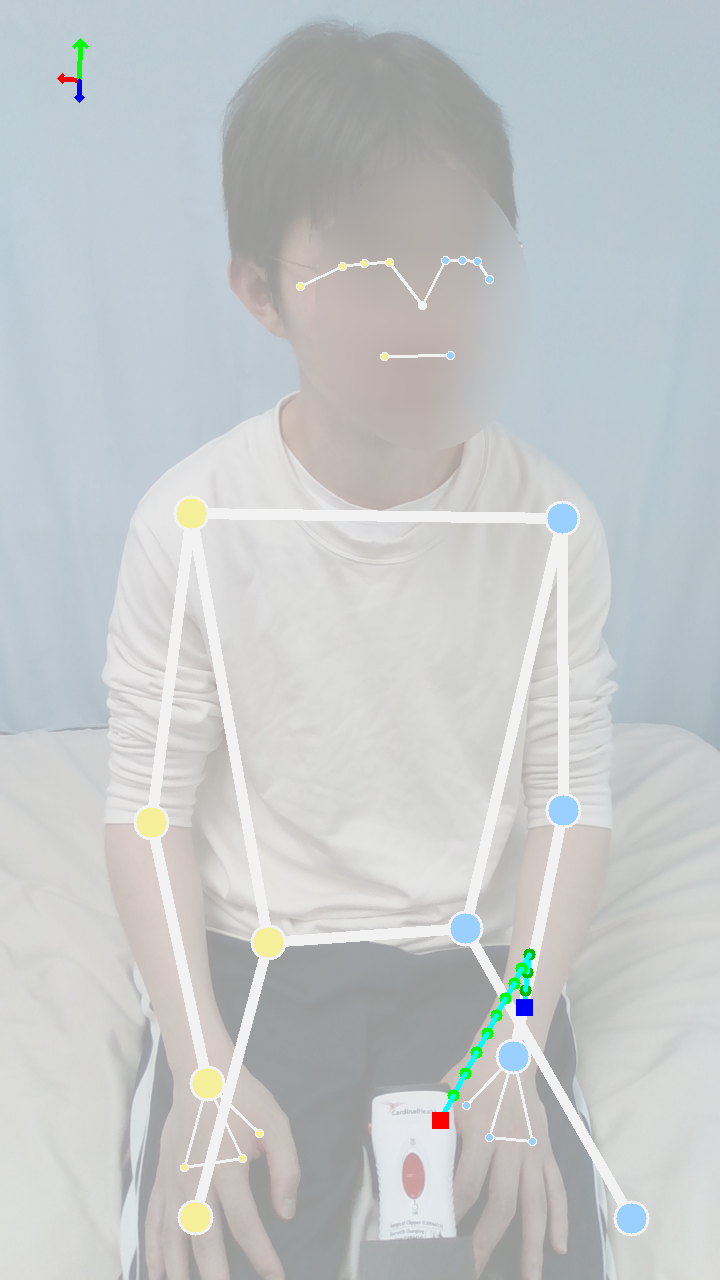}
      \caption{Segment 3}
    \end{subfigure}
    \caption{Example overlay visualization of trajectory 1 in the shaving task, for participant 13.}
    \label{fig:shaving-1}
\end{figure}

\begin{tcolorbox}[title=Ground Truth Paragraphs, breakable,
    colback=ground_truth_frame!10, colframe=ground_truth_frame!50,
    left=2mm, right=2mm,       
    top=1mm, bottom=1mm]
Statement for segment 1: I'm gently and smoothly moving the clipper in a gentle diagonal straight line/curve along your left forearm from near your wrist area toward your elbow, slowing down as I approach near your elbow to lower it and get perfectly lined up, ready before I start a soft grooming action and neat hair removal. You don't need to do anything actively--please just relax and keep your wrist and left arm steady, exposed, and accessible, as I'm not touching you yet.\\

Statement for segment 2: I'm now carefully gliding the clipper in a straight line along your forearm from near your left elbow down to your left wrist, keeping a steady slow speed and soft, constant, gentle pressure to help trim and shave your hair and skin evenly. Please continue to keep your left arm relaxed, steady, and exposed so that I can trim safely in my work.\\

Statement for segment 3: I'm lifting the clipper away from your forearm skin, moving upward and retreating from your left wrist and then quickly diagonally back to my starting position near your left/right wrist area in a straight line, smoothly picking up speed as I go after leaving your forearm. I'm moving in a graceful curved diagonal sweep, and I'm stepping away from your body without touching you anymore. You don't need to do anything now, so you can keep your arms and wrists relaxed and stay still as I safely finish the trimming and move away.
\end{tcolorbox}

\begin{tcolorbox}[title=Summaries from LLM, breakable,
    colback=summary_frame!10, colframe=summary_frame!50,
    left=2mm, right=2mm,       
    top=1mm, bottom=1mm]
Statement for segment 1: I'll move the clipper in a smooth diagonal curve up your left forearm toward your elbow, slowing as I near the elbow to align it for grooming—please relax and keep your wrist and arm steady and exposed.\\

Statement for segment 2: I'll glide the clipper in a straight line down your left forearm toward your wrist at a steady slow speed with constant gentle pressure to trim evenly—please keep your arm relaxed, steady, and exposed.\\

Statement for segment 3: I'll lift the clipper off your skin, then sweep it in a graceful curved diagonal path back to my start position while accelerating away—please remain still with your arms and wrists relaxed as I finish and move away.
\end{tcolorbox}

\begin{tcolorbox}[title=Sample communication generated from CoRI (\textbf{P1}), breakable,
    colback=comm_frame!10, colframe=comm_frame!50,
    left=2mm, right=2mm,       
    top=1mm, bottom=1mm]
Statement for segment 1: I'm moving from your wrist toward your left elbow and slowing down as I approach so I can get set to trim your arm hair—please keep your arm still.\\

Statement for segment 2: I'm now sliding from your left elbow back to your wrist with a steady, gentle pressure to trim your hair, so please continue to keep your arm perfectly still.\\

Statement for segment 3: I'm retreating quickly from your left wrist without touching your skin, so you can relax your arm.
\end{tcolorbox}

\begin{tcolorbox}[title=Sample communication generated from CoRI (\textbf{P7}), breakable,
    colback=comm_frame!10, colframe=comm_frame!50,
    left=2mm, right=2mm,       
    top=1mm, bottom=1mm]
Statement for segment 1: I'm moving the clipper smoothly from near your left wrist up in a gentle diagonal line toward your elbow—slowing down as I near the elbow to get perfectly lined up; you don't need to do anything, just keep your arm relaxed.\\

Statement for segment 2: Now I'm sliding the clipper straight down along your forearm from your elbow to your wrist with a slow, steady touch to groom your skin, so please hold your left arm steady during this part.\\

Statement for segment 3: Finally, I'm lifting the clipper away by moving upward from your wrist along your forearm, starting at a moderate pace and then picking up speed, and you can just continue keeping your arm relaxed.
\end{tcolorbox}

\begin{tcolorbox}[title=Sample communication generated from CoRI (\textbf{P13}), breakable,
    colback=comm_frame!10, colframe=comm_frame!50,
    left=2mm, right=2mm,       
    top=1mm, bottom=1mm]
Statement for segment 1: I'm gently moving the clipper in a smooth diagonal line toward your elbow and then slowing down as I lower it near your elbow to start the neat hair removal—please keep your left arm steady.\\

Statement for segment 2: I'm now carefully sliding the clipper from your elbow to your wrist with a soft, constant pressure for an even shave, so please hold your arm still and in place.\\

Statement for segment 3: I'm swiftly retracting the clipper by moving upward from your wrist and then diagonally back toward me, so you can relax as I finish the grooming process.
\end{tcolorbox}

\textbf{Trajectory 2}

\begin{figure}[bh]
    \centering
    \begin{subfigure}[t]{0.28\textwidth}  
      \centering
      \includegraphics[width=\linewidth]{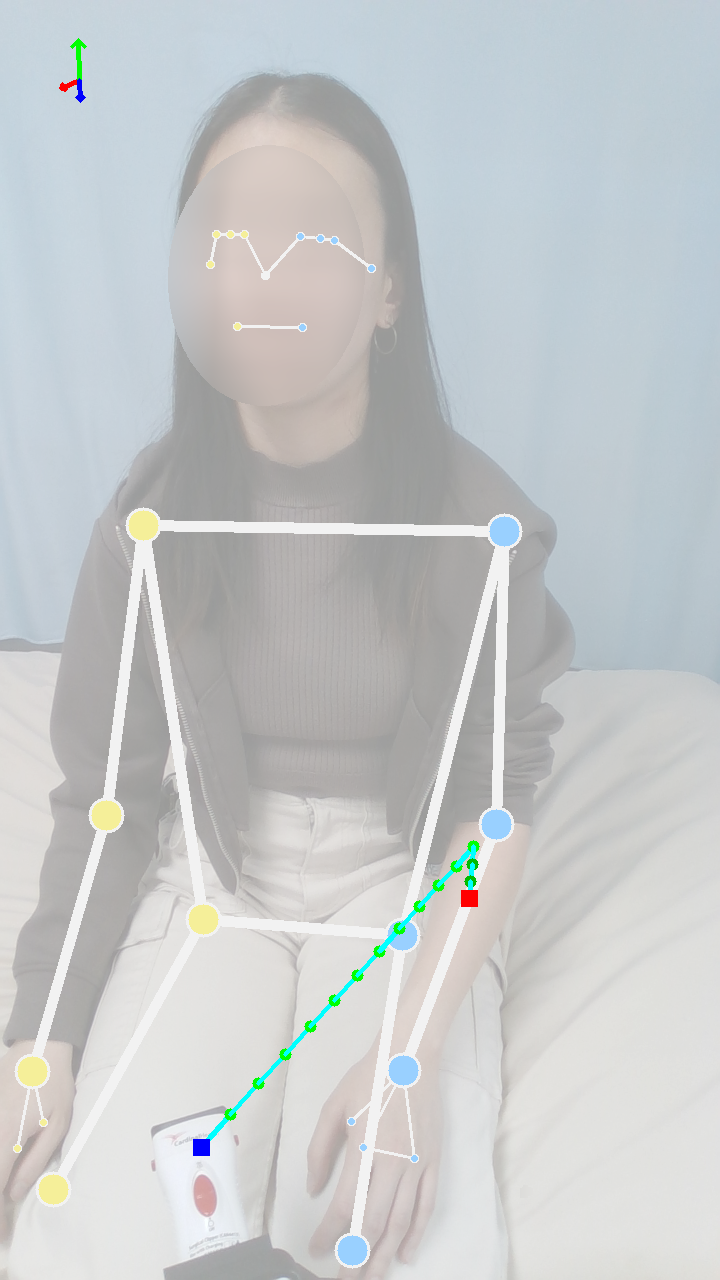}  
      \caption{Segment 1}
    \end{subfigure}\hfill
    \begin{subfigure}[t]{0.28\textwidth}
      \centering
      \includegraphics[width=\linewidth]{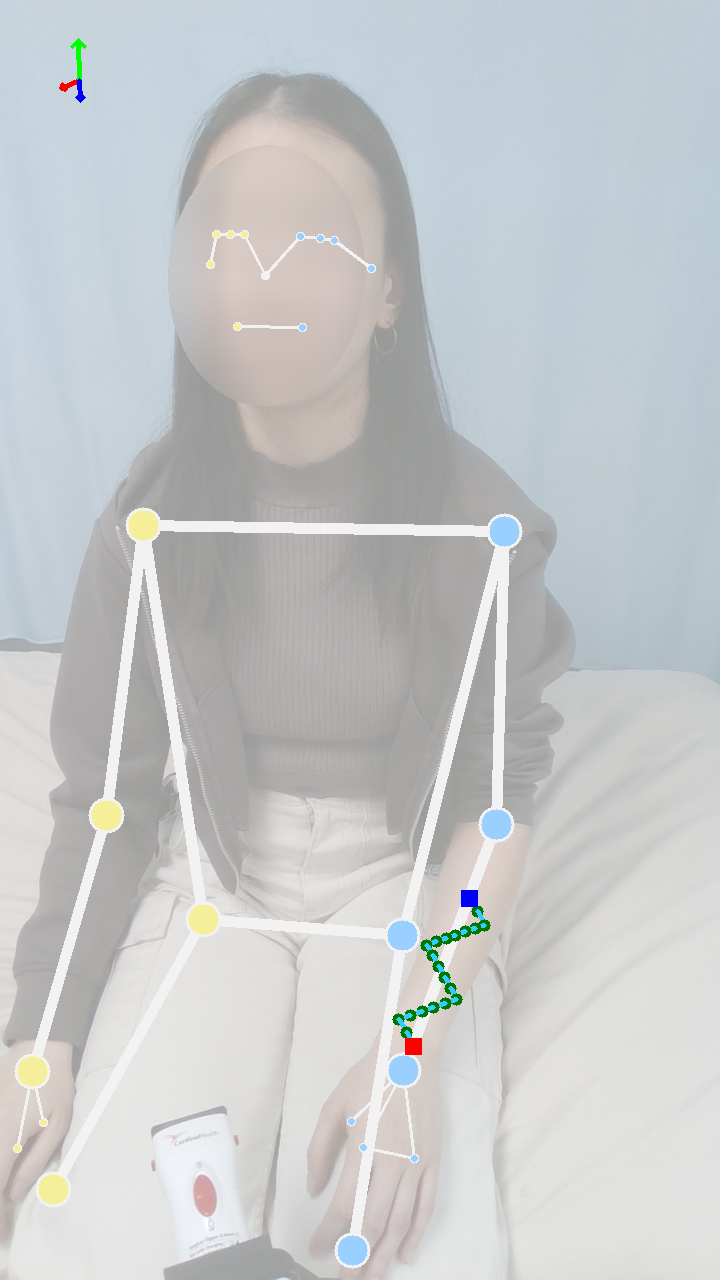}
      \caption{Segment 2}
    \end{subfigure}\hfill
    \begin{subfigure}[t]{0.28\textwidth}
      \centering
      \includegraphics[width=\linewidth]{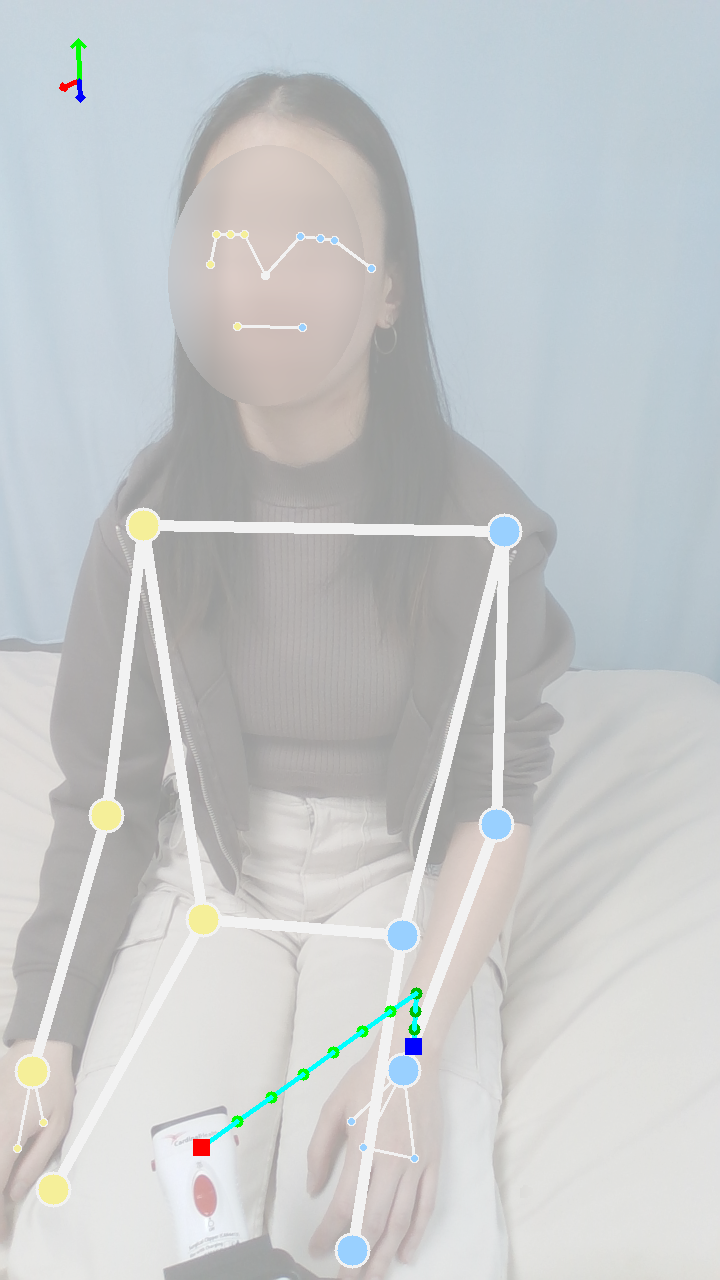}
      \caption{Segment 3}
    \end{subfigure}
    \caption{Example overlay visualization of trajectory 2 in the shaving task, for participant 2.}
    \label{fig:shaving-2}
\end{figure}

\begin{tcolorbox}[title=Ground Truth Paragraphs, breakable,
    colback=ground_truth_frame!10, colframe=ground_truth_frame!50,
    left=2mm, right=2mm,       
    top=1mm, bottom=1mm]
Statement for segment 1: I'm gently and smoothly moving the clipper in a gentle diagonal straight line/curve up along your left forearm from near your left/right wrist area toward your elbow, starting at a moderate speed and slowing down as I approach close to your elbow to lower it and carefully get perfectly lined up, ready before I start a soft grooming action and neat hair removal. You don't need to do anything actively--please just relax and keep your wrist and left arm steady, exposed, and accessible so that I can reach the area, as I'm not touching you yet, and just getting into the perfect position, setting up to help with gentle clipping on your left forearm.\\

Statement for segment 2: I'm now carefully gliding the clipper in a fun zigzag path along your forearm from near your left elbow down to your left wrist, keeping a steady slow speed and soft, constant, gentle pressure to help trim and shave your hair and skin evenly. Please continue to keep your left arm relaxed, steady, and exposed so that I can trim safely in my work.\\

Statement for segment 3: I'm lifting the clipper upward and away from your left wrist, and then diagonally back to my starting position in the open space near your left/right wrist area in a straight line, starting at a gentle pace and smoothly picking up speed as I go faster after leaving your forearm, without applying any extra pressure or touching your skin anymore so that I'm clear of your arm. You don't need to do anything now, so you can keep your arms and wrists relaxed and stay still as I safely finish the trimming. The grooming is now complete.
\end{tcolorbox}

\begin{tcolorbox}[title=Summaries from LLM, breakable,
    colback=summary_frame!10, colframe=summary_frame!50,
    left=2mm, right=2mm,       
    top=1mm, bottom=1mm]
Statement for segment 1: I'll move the clipper diagonally up to your left elbow, gradually slowing as I approach to align it for grooming—please keep your left arm relaxed, steady, and exposed.\\

Statement for segment 2: I'll glide the clipper in a zigzag pattern down to your left wrist with constant gentle pressure to trim your hair evenly—please keep your left arm relaxed, steady, and exposed.\\

Statement for segment 3: I'll lift the clipper upward away from your left wrist then diagonally return it to a rest position near your wrist while accelerating to complete the trim—please keep your arms and wrists relaxed and still.
\end{tcolorbox}

\begin{tcolorbox}[title=Sample communication generated from CoRI (\textbf{P2}), breakable,
    colback=comm_frame!10, colframe=comm_frame!50,
    left=2mm, right=2mm,       
    top=1mm, bottom=1mm]
Statement for segment 1: I'm moving in a straight line toward your left elbow, gradually slowing down as I approach to gently get ready for grooming; please keep your arm steady so I can do this safely.\\

Statement for segment 2: I'm now following a friendly zigzag path along your left forearm, applying a gentle and steady pressure as I smooth it out—please continue to hold your arm still.\\

Statement for segment 3: I'm returning in a straight line from your left wrist back to my starting spot, gradually picking up speed, and I won't be touching your arm now, so you can relax.
\end{tcolorbox}

\begin{tcolorbox}[title=Sample communication generated from CoRI (\textbf{P5}), breakable,
    colback=comm_frame!10, colframe=comm_frame!50,
    left=2mm, right=2mm,       
    top=1mm, bottom=1mm]
Statement for segment 1: I'm gently moving the clipper from near your wrist upward and outward along your forearm toward your elbow, slowing down as I get close for careful alignment—please keep your left arm still so I can set up for trimming.\\

Statement for segment 2: Now I'm gliding the clipper down toward your wrist along a fun zig-zag path while keeping a gentle, steady touch on your skin—please keep your forearm steady so I can trim safely.\\

Statement for segment 3: Finally, I'm lifting the clipper a bit and then quickly moving it away from your wrist and arm without touching you, so feel free to relax your arm as I finish up.
\end{tcolorbox}

\begin{tcolorbox}[title=Sample communication generated from CoRI (\textbf{P12}), breakable,
    colback=comm_frame!10, colframe=comm_frame!50,
    left=2mm, right=2mm,       
    top=1mm, bottom=1mm]
Statement for segment 1: I'm moving the clipper in a smooth, diagonal line along your left forearm toward your elbow—starting at a moderate speed and then slowing down as I get closer—to get ready for a gentle trim, so please just relax your arm.\\

Statement for segment 2: Now, I'm following a gentle, winding path from your elbow down to your wrist, applying a steady, light pressure as I trim your hair; please keep your left arm still and exposed so I can work safely.\\

Statement for segment 3: Finally, I'm lifting the clipper upward from your wrist and then quickly moving it away into open space, so just stay comfortable as I finish up.
\end{tcolorbox}

\subsection{Feeding}

\textbf{Trajectory 1}

\begin{figure}[tbh]
    \centering
    \begin{subfigure}[t]{0.48\textwidth}  
      \centering
      \includegraphics[width=\linewidth]{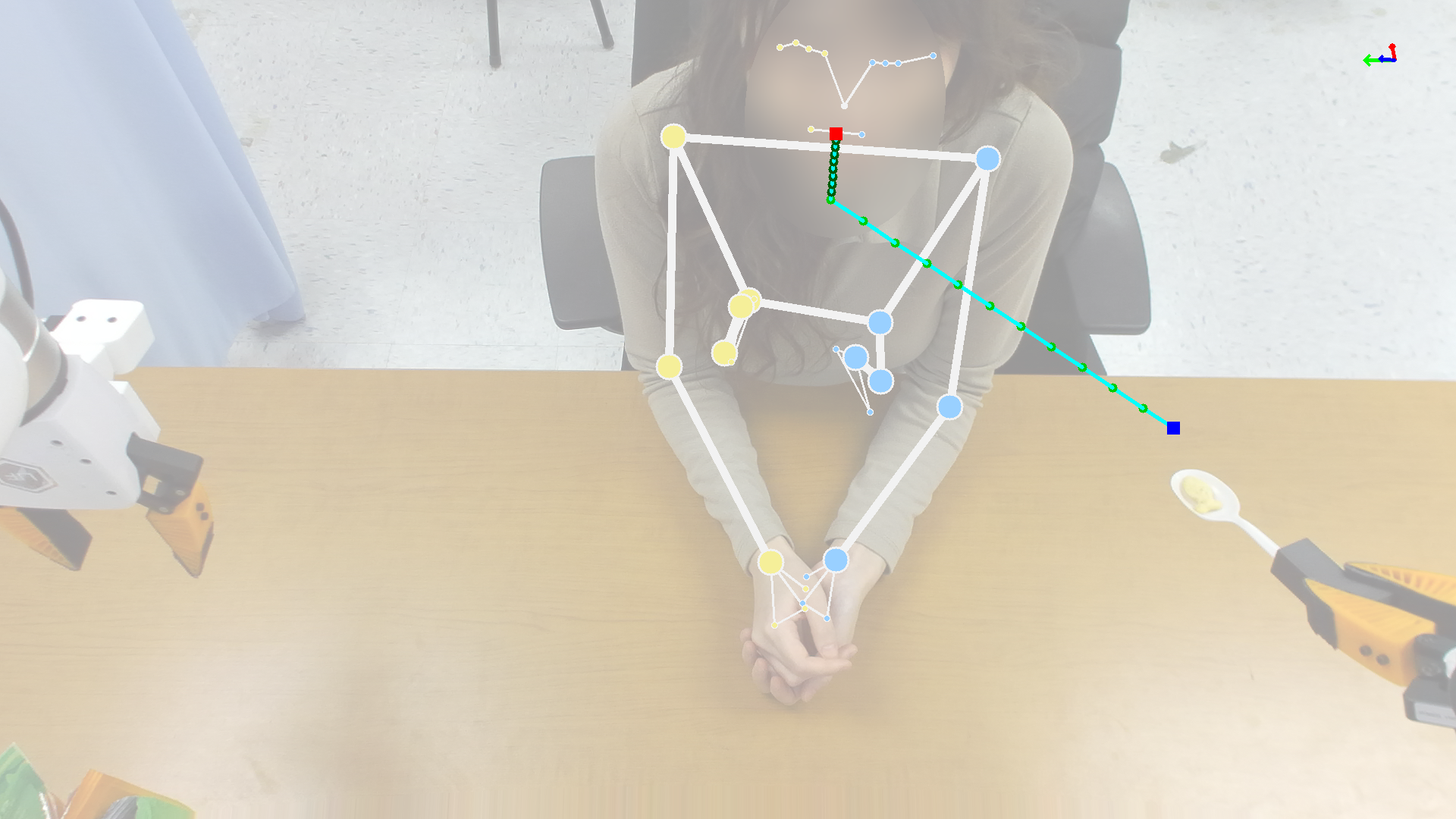}  
      \caption{Segment 1}
    \end{subfigure}\hfill
    \begin{subfigure}[t]{0.48\textwidth}
      \centering
      \includegraphics[width=\linewidth]{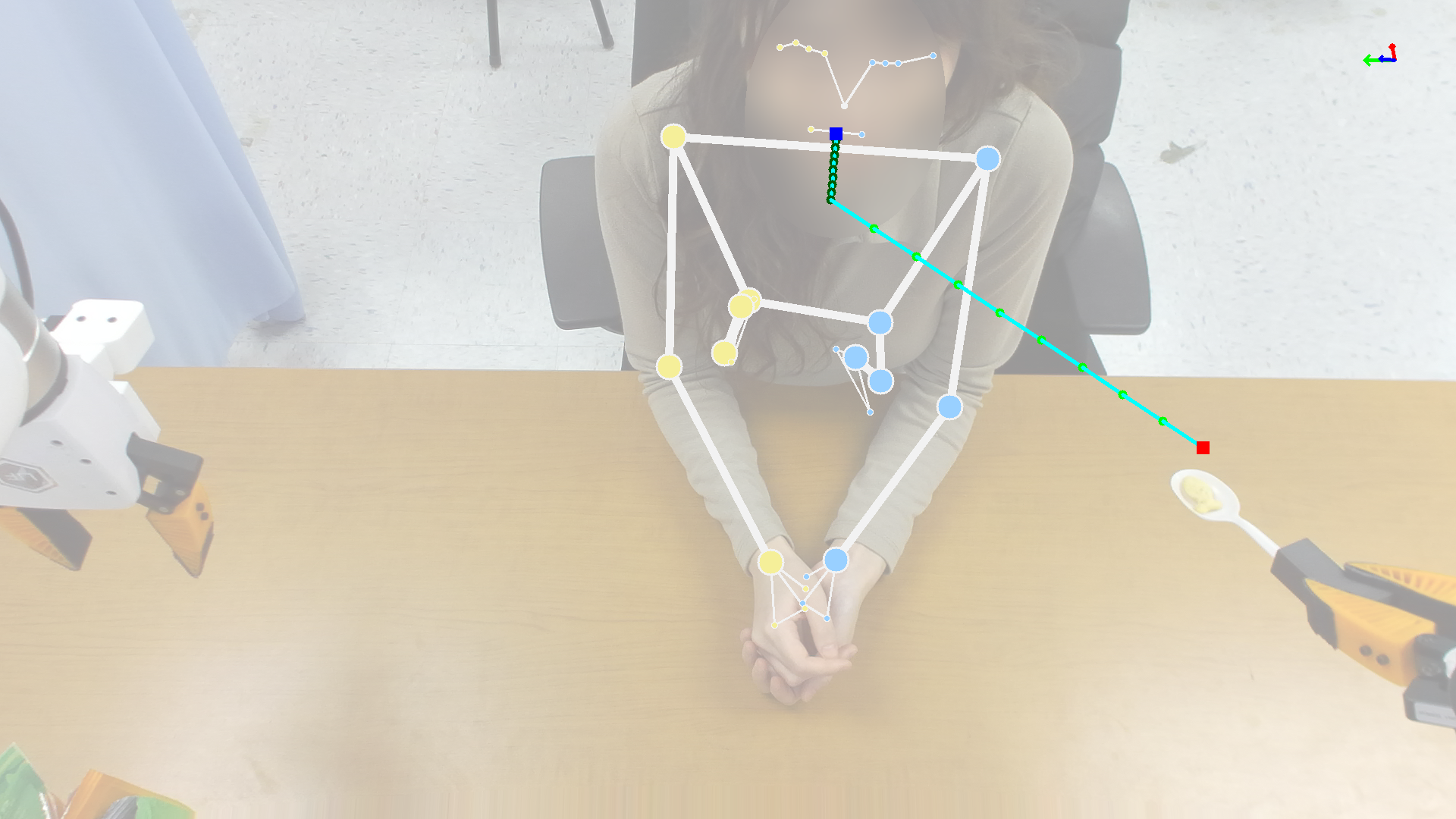}
      \caption{Segment 2}
    \end{subfigure}
    \caption{Example overlay visualization of trajectory 1 in the feeding task, for participant 11.}
    \label{fig:feeding-1}
\end{figure}

\begin{tcolorbox}[title=Ground Truth Paragraphs, breakable,
    colback=ground_truth_frame!10, colframe=ground_truth_frame!50,
    left=2mm, right=2mm,       
    top=1mm, bottom=1mm]
Statement for segment 1: I'm moving the spoon with food, likely some delicious cracker, in a straight line diagonally toward your mouth at a steady pace and slowing down into a straight vertical approach gently as I get close. Please be ready to open your mouth a bit, take the food, and enjoy your bite when it arrives and I pause a bit, to join in our friendly meal.\\

Statement for segment 2: Please take the food from the spoon during my pause. Then I'm smoothly and gently pulling the spoon away from your mouth in a straight line and then speed up to retract it diagonally back to my spot on the table, so just relax and enjoy your meal, and you don't need to do anything as I finish our little sharing moment.
\end{tcolorbox}

\begin{tcolorbox}[title=Summaries from LLM, breakable,
    colback=summary_frame!10, colframe=summary_frame!50,
    left=2mm, right=2mm,       
    top=1mm, bottom=1mm]
Statement for segment 1: I'll move the spoon holding the food diagonally toward your mouth at a steady pace, then slow into a vertical approach as I get close—pause for you to open your mouth and take the bite.\\

Statement for segment 2: After you take the food during my pause, I'll smoothly pull the spoon straight back then speed up as I retract it diagonally to my spot on the table, so you can just relax and enjoy without doing anything further.
\end{tcolorbox}

\begin{tcolorbox}[title=Sample communication generated from CoRI (\textbf{P4}), breakable,
    colback=comm_frame!10, colframe=comm_frame!50,
    left=2mm, right=2mm,       
    top=1mm, bottom=1mm]
Statement for segment 1: I'm moving your spoon up to your mouth using a quick diagonal path that slows into a near-vertical approach--please open your mouth to let you receive your food.\\

Statement for segment 2: I'm now gently retracting the spoon away from your mouth along a smooth path and then quickly returning it to the table edge; you don't need to do anything during this move.
\end{tcolorbox}

\begin{tcolorbox}[title=Sample communication generated from CoRI (\textbf{P10}), breakable,
    colback=comm_frame!10, colframe=comm_frame!50,
    left=2mm, right=2mm,       
    top=1mm, bottom=1mm]
Statement for segment 1: I'm moving the spoon straight toward your mouth, slowing down as I get closer, so please open your mouth to take the food.\\

Statement for segment 2: I'm now gently pulling the spoon away from your mouth—starting slowly and then speeding up as I return to the table—and you don't need to do anything.
\end{tcolorbox}

\begin{tcolorbox}[title=Sample communication generated from CoRI (\textbf{P11}), breakable,
    colback=comm_frame!10, colframe=comm_frame!50,
    left=2mm, right=2mm,       
    top=1mm, bottom=1mm]
Statement for segment 1: I'm moving my spoon in a straight line toward your mouth and slowing down as I get close—please be ready to take the food when it arrives.  \\

Statement for segment 2: I'm smoothly pulling the spoon away from your mouth in a straight line and then retracting it diagonally back to my spot, so just relax as I move.
\end{tcolorbox}

\textbf{Trajectory 2}

\begin{figure}[tbh]
    \centering
    \begin{subfigure}[t]{0.48\textwidth}  
      \centering
      \includegraphics[width=\linewidth]{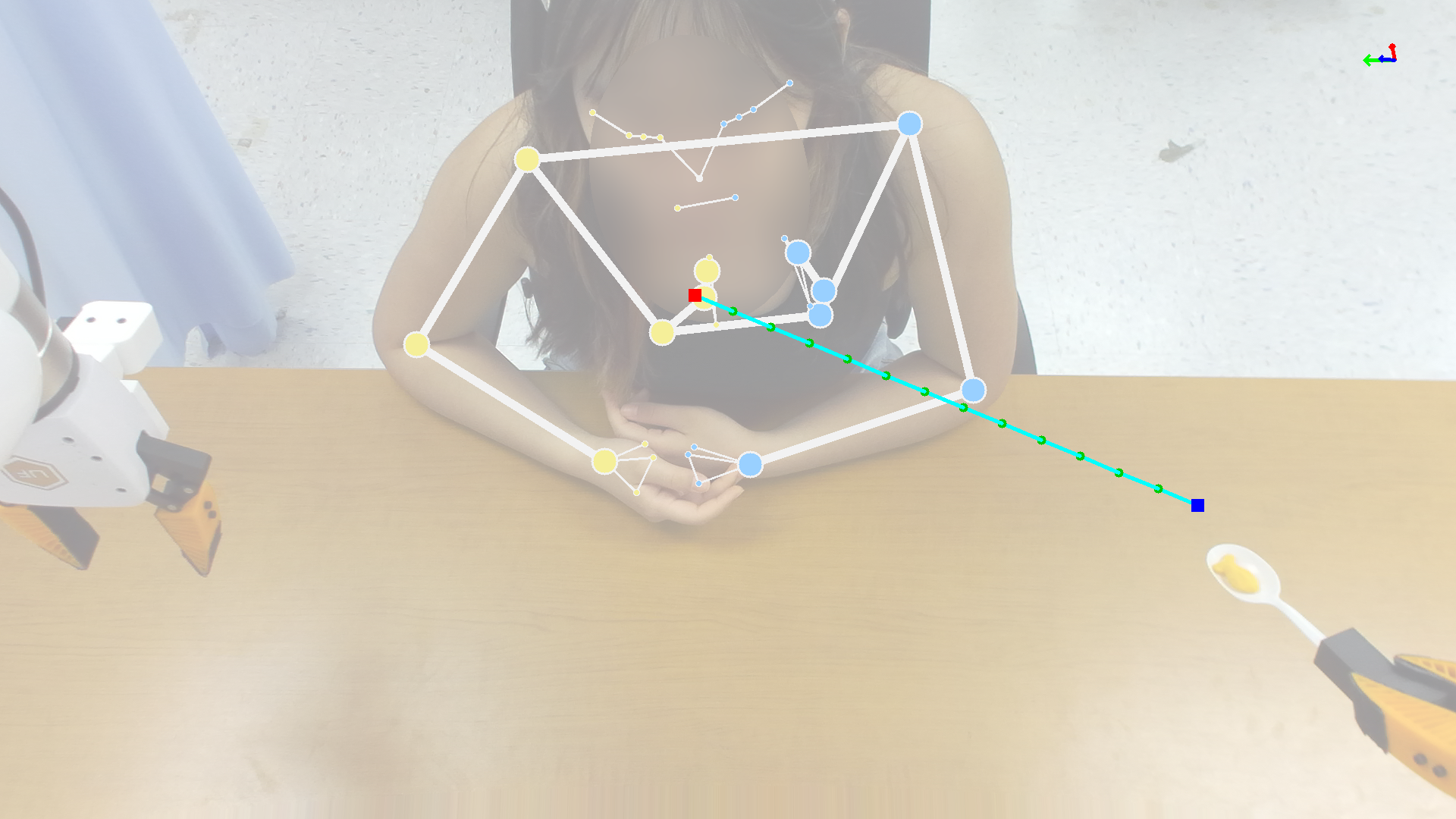}  
      \caption{Segment 1}
    \end{subfigure}\hfill
    \begin{subfigure}[t]{0.48\textwidth}
      \centering
      \includegraphics[width=\linewidth]{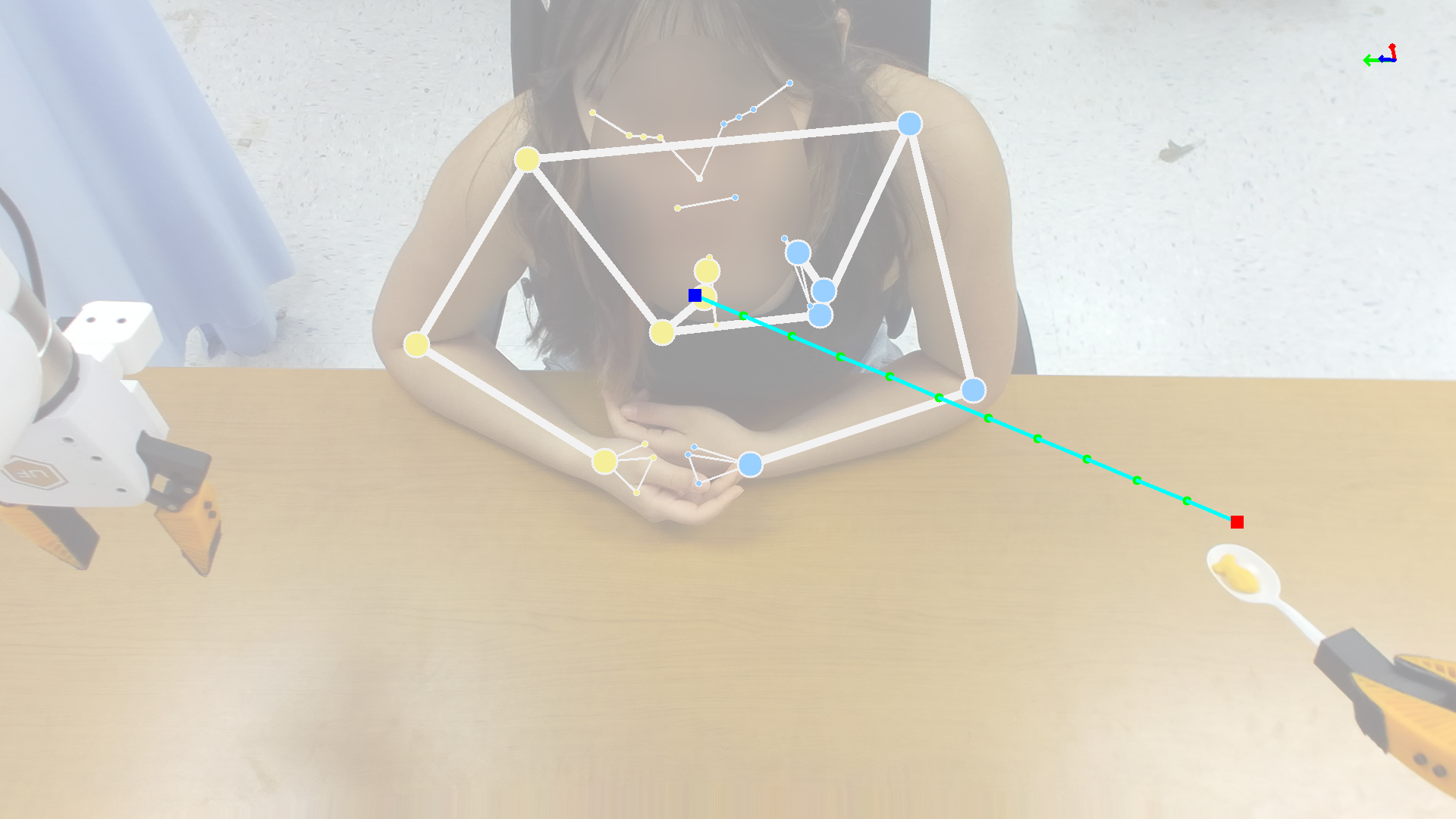}
      \caption{Segment 2}
    \end{subfigure}
    \caption{Example overlay visualization of trajectory 2 in the feeding task, for participant 16.}
    \label{fig:feeding-2}
\end{figure}

\begin{tcolorbox}[title=Ground Truth Paragraphs, breakable,
    colback=ground_truth_frame!10, colframe=ground_truth_frame!50,
    left=2mm, right=2mm,       
    top=1mm, bottom=1mm]
Statement for segment 1: I'm kindly moving the spoon with tasty food, likely some delicious cracker, in a straight line diagonally toward just in front of your mouth at a steady pace to help you start your meal. There's nothing you need to do at first, so just sit back and relax, but please be ready to lean forward, open your mouth and take the food when it arrives and I get close and pause a bit to offer you the bite, so you can enjoy your treat.\\

Statement for segment 2: I'm holding the spoon by your mouth for a moment of pause so that you have plenty of time to take the food and savor your snack. Please lean forward to take a bite with your mouth or use your hand to take the food. Then I'm smoothly and gently pulling the spoon away from your mouth in a straight line faster, retracting it diagonally back to my spot on the table, so just relax, and you don't need to do anything.
\end{tcolorbox}

\begin{tcolorbox}[title=Summaries from LLM, breakable,
    colback=summary_frame!10, colframe=summary_frame!50,
    left=2mm, right=2mm,       
    top=1mm, bottom=1mm]
Statement for segment 1: I'll move the spoon diagonally to just in front of your mouth at a steady pace to offer you a bite—please lean forward, open your mouth, and take the food when I pause.\\

Statement for segment 2: I'll hold the spoon by your mouth so you can lean forward or use your hand to take a bite, then smoothly retract it diagonally back to the table at a faster speed.
\end{tcolorbox}

\begin{tcolorbox}[title=Sample communication generated from CoRI (\textbf{P1}), breakable,
    colback=comm_frame!10, colframe=comm_frame!50,
    left=2mm, right=2mm,       
    top=1mm, bottom=1mm]
Statement for segment 1: I'm moving straight toward your mouth with the spoon holding your snack—please open your mouth a little so you can receive your treat.\\

Statement for segment 2: Now I'm gently retracting the spoon in a straight line away from your mouth; you can just relax as I pull back.
\end{tcolorbox}

\begin{tcolorbox}[title=Sample communication generated from CoRI (\textbf{P8}), breakable,
    colback=comm_frame!10, colframe=comm_frame!50,
    left=2mm, right=2mm,       
    top=1mm, bottom=1mm]
Statement for segment 1: I'm moving in a straight line to bring this spoonful of yummy food right near your mouth—please open your mouth when you see it coming so you can take a bite. \\

Statement for segment 2: I'm moving straight back from your mouth, gently returning the spoon to the table, and you don't need to do anything now.
\end{tcolorbox}

\begin{tcolorbox}[title=Sample communication generated from CoRI (\textbf{P16}), breakable,
    colback=comm_frame!10, colframe=comm_frame!50,
    left=2mm, right=2mm,       
    top=1mm, bottom=1mm]
Statement for segment 1: I'm moving straight toward your mouth with a yummy cracker on a spoon, so please open your mouth to receive your treat when I arrive.  \\

Statement for segment 2: I'm now gently retracting the spoon away from your mouth along a direct path, so please remain still as I complete my move.
\end{tcolorbox}
\end{document}